\PassOptionsToPackage{table}{xcolor}
\documentclass{article}

\PassOptionsToPackage{numbers, compress}{natbib}

\usepackage[preprint]{neurips_2025}

\usepackage[utf8]{inputenc} 
\usepackage[T1]{fontenc}    
\usepackage{hyperref}       
\usepackage{url}            
\usepackage{booktabs}       
\usepackage{amsfonts}       
\usepackage{nicefrac}       
\usepackage{microtype}      
\usepackage{xcolor}         
\usepackage{multirow}
\usepackage{float}
\usepackage{amsmath}
\usepackage{subcaption}
\usepackage{wrapfig}
\usepackage[textsize=tiny]{todonotes}
\usepackage[most]{tcolorbox}

\newtcolorbox{prompt}[1]{
    enhanced,
    left=4mm,
    right=4mm,
    top=2mm,
    bottom=2mm,
    boxsep=0mm,
    rounded corners,
    title=#1,
    fontupper=\footnotesize\linespread{0.9}\fontfamily{lmr}\selectfont,
    }

\definecolor{darkgreen}{rgb}{0, 0.5001960, 0}
\definecolor{darkred}{rgb}{0.8, 0, 0}
\definecolor{darkblue}{rgb}{0, 0, 0.545}

\definecolor{mygreen1}{rgb}{0.9686, 0.9804, 0.9843}  
\definecolor{mygreen2}{rgb}{0.7961, 0.8745, 0.9216}  
\definecolor{mygreen3}{rgb}{0.6235, 0.7647, 0.8588}  
\definecolor{mygreen4}{rgb}{0.3686, 0.5961, 0.7608} 

\title{Learning from Peers in Reasoning Models}

\author{Tongxu Luo$^1$\thanks{~~Equal Contributions.}
\quad Wenyu Du$^{2*}$
\quad Jiaxi Bi$^3$ \quad Stephen Chung$^{2}$ \quad Zhengyang Tang$^1$ \\
\bf{Hao Yang$^4$ \quad Min Zhang$^4$ \quad Benyou Wang$^1$\thanks{~~Corresponding Author.}}\\
$^1$The Chinese University of Hong Kong, Shenzhen \\
$^2$DualityRL \quad $^3$USTB \quad $^4$Huawei \\
\texttt{tongxuluo@gmail.com \quad wenyu.du@dualityrl.com \quad wangbenyou@cuhk.edu.cn}
}

\begin{document}
\maketitle

\begin{abstract}
Large Reasoning Models (LRMs) have the ability to self-correct even when they make mistakes in their reasoning paths.
However, our study reveals that when the reasoning process starts with a short but poor beginning, it becomes difficult for the model to recover.
We refer to this phenomenon as the \emph{``Prefix Dominance Trap''}.
Inspired by psychological findings that peer interaction can promote self-correction without negatively impacting already accurate individuals, we propose \textbf{Learning from Peers}~(LeaP) to address this phenomenon.
Specifically, every \( T \) tokens, each reasoning path summarizes its intermediate reasoning and shares it with others through a routing mechanism, enabling paths to incorporate peer insights during inference.
However, we observe that smaller models sometimes fail to follow summarization and reflection instructions effectively.
To address this, we fine-tune them into our \textbf{LeaP-T} model series.
Experiments on AIME 2024, AIME 2025, AIMO 2025, and GPQA Diamond show that LeaP provides substantial improvements.
For instance, QwQ-32B with LeaP achieves nearly 5 absolute points higher than the baseline on average, and surpasses DeepSeek-R1-671B on three math benchmarks with an average gain of 3.3 points.
Notably, our fine-tuned LeaP-T-7B matches the performance of DeepSeek-R1-Distill-Qwen-14B on AIME 2024.
In-depth analysis reveals LeaP's robust error correction by timely peer insights, showing strong error tolerance and handling varied task difficulty.
LeaP marks a milestone by enabling LRMs to collaborate during reasoning.
Our code, datasets, and models are available at \url{https://learning-from-peers.github.io/}.
\end{abstract}

\section{Introduction}

Large reasoning models (LRMs)~\cite{openai2024a,openai2024b,openai2025o3,guo2025deepseek,qwq32b} demonstrate strong performance on complex reasoning tasks. 
This success is largely attributed to their self-correction capability in test-time scaling~\cite{zhang2025and,snell2024scaling,zeng2025revisiting,chen2024expanding}, which consists of two emergent features: during generation, LRMs evaluate their current reasoning trajectories (self-verification) and may generate alternative ones (self-refinement) — a behavior often referred to as the ``aha moment''~\cite{guo2025deepseek}. 
However, recent research reveals notable limitations in this self-correction mechanism. 
For example, researchers~\cite{zeng2025revisiting} observe that LRMs frequently become stuck in incorrect reasoning paths that are rarely corrected. 
This raises an important question: 
\textbf{Is the self-correction ability of LRMs truly as strong as we think?}

To answer this question, we design a task where LRMs are required to solve problems starting from fixed beginnings, extracted from responses that lead to either correct or incorrect answers.
If an LRM possesses strong self-correction ability, it should be able to recover from a low-quality beginning and still arrive at the correct answer.
Surprisingly, many LRMs, including QwQ-32B~\cite{qwq32b}, exhibit a performance drop of nearly 20\% when starting from a poor beginning that accounts for only 15\% of the average response length.
This striking result suggests that \textbf{the self-correction ability of LRMs is significantly weaker than previously thought}.
A short and flawed prelude can lead to a substantial degradation in final performance.
We refer to this phenomenon as the \emph{``Prefix Dominance Trap''}.

Recent findings~\cite{giuliodori2006peer,snyder2015implementation,falk2006clean} in psychology indicate that peer-based instruction effectively supports students in correcting misconceptions and improving performance with minimal negative impact.
Inspired by this, we hypothesize that enabling LRMs to engage in peer learning may extend their self-verification capabilities beyond their own reasoning to include the reasoning of others.
This broadens the search space for refinement, allowing LRMs to focus more on verifying existing reasoning paths rather than generating new ones from scratch.
Since verification is generally less complex than generation, this shift reduces the cognitive burden and facilitates more effective reasoning refinement.
Building on this insight, we propose \textbf{Learning from Peers}~(LeaP) to improve reasoning in large models: instead of relying solely on individual reasoning paths, \textbf{we enable LRMs to conduct cross-path interaction during parallel inference.}
Concretely, during generation, after every \( T \) tokens, each reasoning path summarizes its intermediate reasoning into a concise message, which is then shared with other paths through a heuristic routing mechanism.
Simultaneously, each path receives summaries from its peers.

We first validate our hypothesis by placing LeaP under the \emph{``Prefix Dominance Trap''} setting. Experimental results show that models with LeaP reduce the performance gap by nearly 10\% compared to those without LeaP.
This finding suggests that LeaP encourages each reasoning path to verify not only its own trajectory but also those of its peers, thereby decreasing the risk of overlooking correct solutions.
To further assess the effectiveness of LeaP, we conduct comprehensive evaluations on AIME 2024~\cite{aime2024}, AIME 2025~\cite{aime2025}, the reference set of AIMO 2025~\cite{ai-mathematical-olympiad-progress-prize-2}, and GPQA Diamond~\cite{rein2024gpqa}.
All reasoning models show significant improvement when using LeaP compared to those without LeaP, under comparable inference token budgets.

However, during these experiments, we observe that smaller models without further training, such as DeepSeek-R1-Distill-Qwen-7B~\cite{guo2025deepseek}, sometimes struggle to follow summarization and reflection instructions effectively~(Case studies in Appendix~\ref{app:case_sutdy_failure}).
To address this, we propose \textbf{LeaP-T model series, including 1.5B, 7B, and 14B, which empowers LeaP with further training adaptation.}
Our experiments show significant improvements.
For example, the LeaP-T-7B model achieves a Pass@1 score of 64.38 on AIME 2024, which is comparable to DeepSeek-R1-Distill-Qwen-14B~\cite{guo2025deepseek} at 64.47.

To summarize, our contributions are as follows:
\begin{itemize}
    \item \textbf{Quantifying the \emph{Prefix Dominance Trap}:} We quantitatively validate the existence of the \emph{Prefix Dominance Trap} in Large Reasoning Models (LRMs), showing that even short low-quality prefixes can lead to significant performance degradation.
    
    \item \textbf{A parallel inference method with cross-path interaction:} We propose LeaP, which enables reasoning paths to communicate during inference. With LeaP, QwQ-32B gains nearly 5 points on average over the baseline and even surpasses DeepSeek-R1-671B~\cite{guo2025deepseek} on various benchmarks. We also conduct a comprehensive analysis of LeaP from multiple perspectives, including sensitivity to communication and robustness under varying conditions.

    \item \textbf{A series of trained models:} We train a series of models that are adapted to the LeaP framework and release them as open-source resources to facilitate future research.
\end{itemize}

\vspace{-1mm}
\section{Enabling Cross-path Interaction in Parallel Inference: LeaP}
\vspace{-1mm}

\subsection{Motivation: Prefix Dominance Trap}\label{sec:motivation}

\begin{wrapfigure}{R}{0.4\textwidth}
    \centering \vspace{-4mm}
    \includegraphics[width=\linewidth]{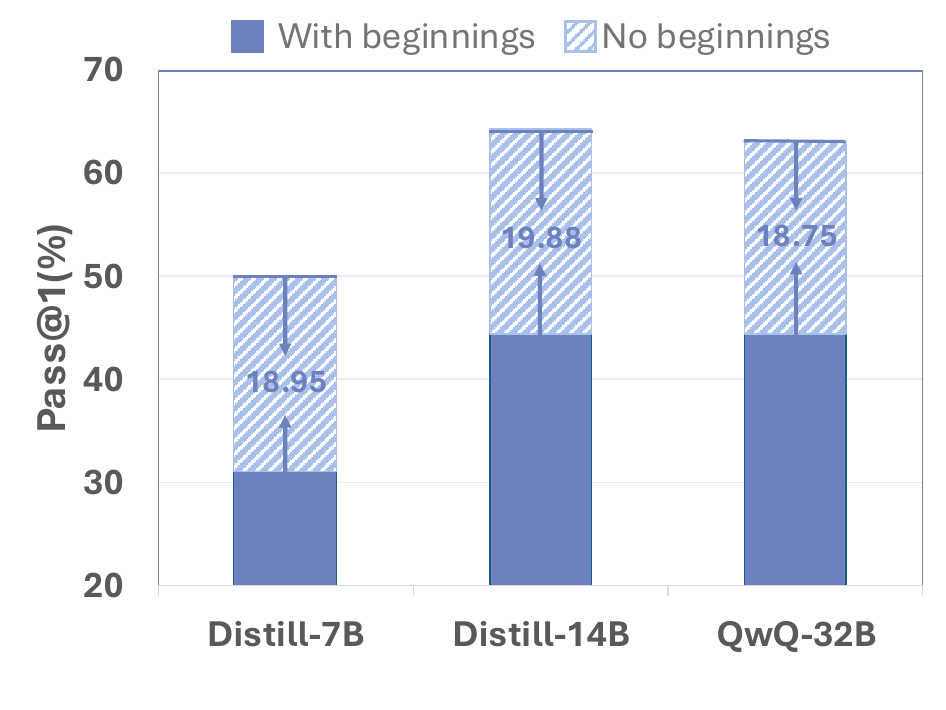}
    \caption{The results of starting with bad beginnings.
    }
    \label{fig:motivation} \vspace{-6mm}
\end{wrapfigure}
To assess the self-correction of LRMs, we introduce a task where LRMs are required to solve problems starting from fixed beginnings.
If an LRM possesses strong correction abilities, it should more easily recover from reasoning paths that may lead to incorrect answers, thereby generating correct responses and resulting in a narrower accuracy gap with no-beginning.

Specifically, we utilize DeepSeek-R1-Distill-Qwen series~\cite{guo2025deepseek} and QwQ-32B~\cite{qwq32b} to generate 32 responses per question in AIME 2024~\cite{aime2024}.
For each model, we randomly select 10 incorrect responses from 10 distinct questions containing both correct and incorrect outputs.
We retain the first $15\%$ of tokens from each selected response as the prefix.
We then generate 16 continuations for each prefix using the original LRMs with temperature $\tau = 1$ to encourage diversity.
To assess LRM self-correction, we compute the average accuracy gap $P_G$ between these constrained and unconstrained generations.
As shown in Figure~\ref{fig:motivation}, we report Pass@1 performance on AIME 2024 subsets (different for each model).
Surprisingly, all selected LRMs exhibit a substantial performance drop of nearly $20\%$ when the first $15\%$ tokens are fixed from incorrect responses.
This result highlights that \textbf{LRM self-correction ability is limited}.
We term this phenomenon the \emph{``Prefix Dominance Trap''}.

\vspace{-1mm}
\subsection{Methodology: Cross-path Interaction in Parallel Inference}\label{sec:method}\vspace{-1mm}

To ameliorate \emph{``Prefix Dominance Trap''}, we introduce Learning from Peers (LeaP), a novel inference-time strategy that draws inspiration from collaborative learning paradigms observed in psychology~\cite{giuliodori2006peer,snyder2015implementation,falk2006clean}.
Our central hypothesis is that enabling LRMs to engage in structured peer interaction can extend their self-verification capabilities.
Instead of solely relying on individual trial-and-error, paths can leverage the reasoning processes of others.
This paradigm broadens the search space for refinement, crucially shifting the LRM's focus from the demanding task of path generation towards the often less complex task of verifying and refining existing diverse paths.
This shift aims to reduce the cognitive burden and enhance the overall effectiveness of reasoning refinement.

\begin{figure}[H]
    \centering \vspace{-2mm}
    \includegraphics[width=0.9\linewidth]{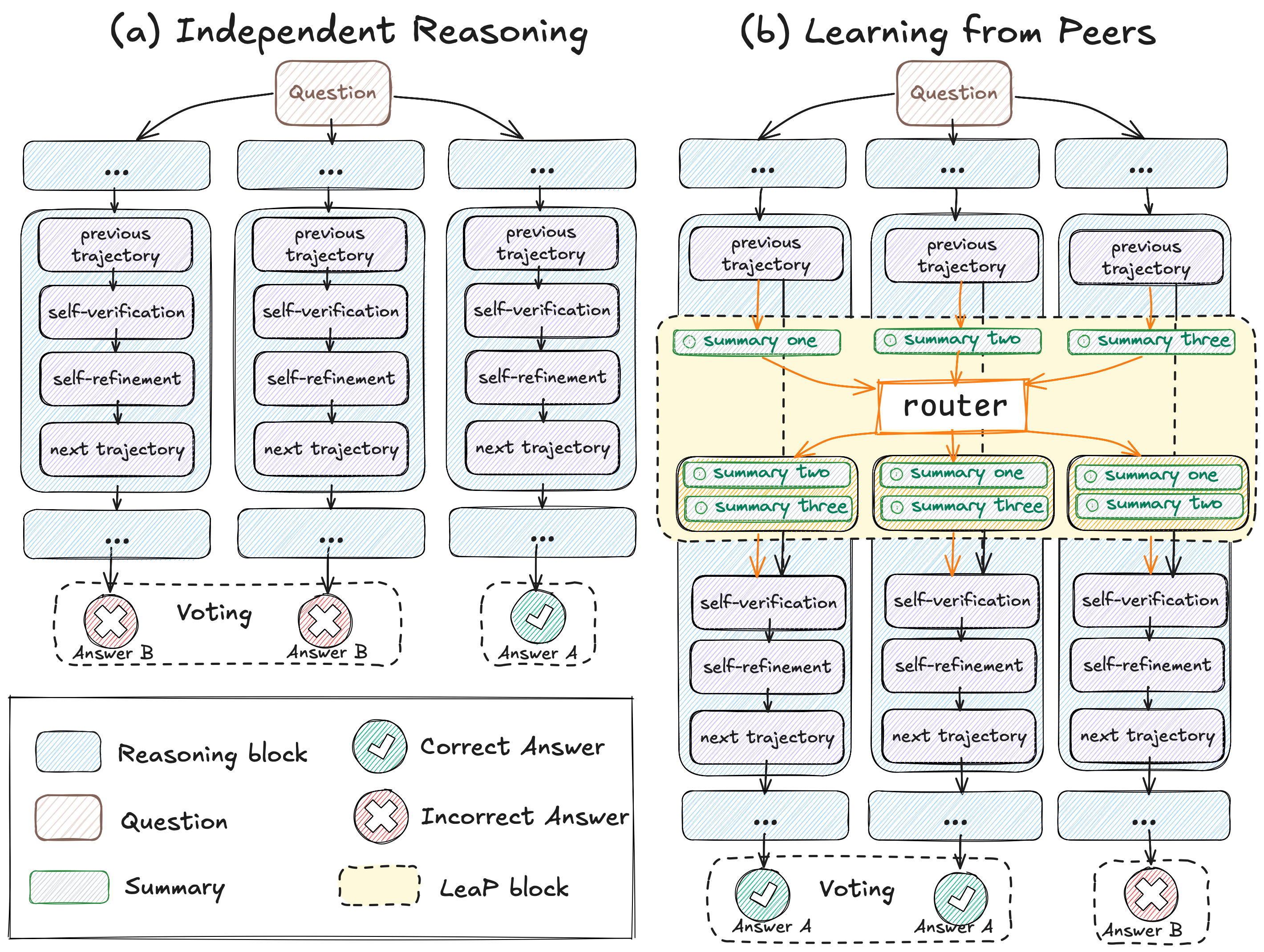}
    \caption{The illustration of (a) Independent Reasoning and (b) the proposed method Learning from Peers (LeaP). In independent reasoning, multiple paths are generated independently in parallel. In contrast, LeaP inserts a LeaP block into reasoning path, encouraging the model to learn from peers.}
    \label{fig:main} \vspace{-3mm}
\end{figure}

\begin{figure}[H]
    \centering \vspace{-3mm}
    \includegraphics[width=\linewidth]{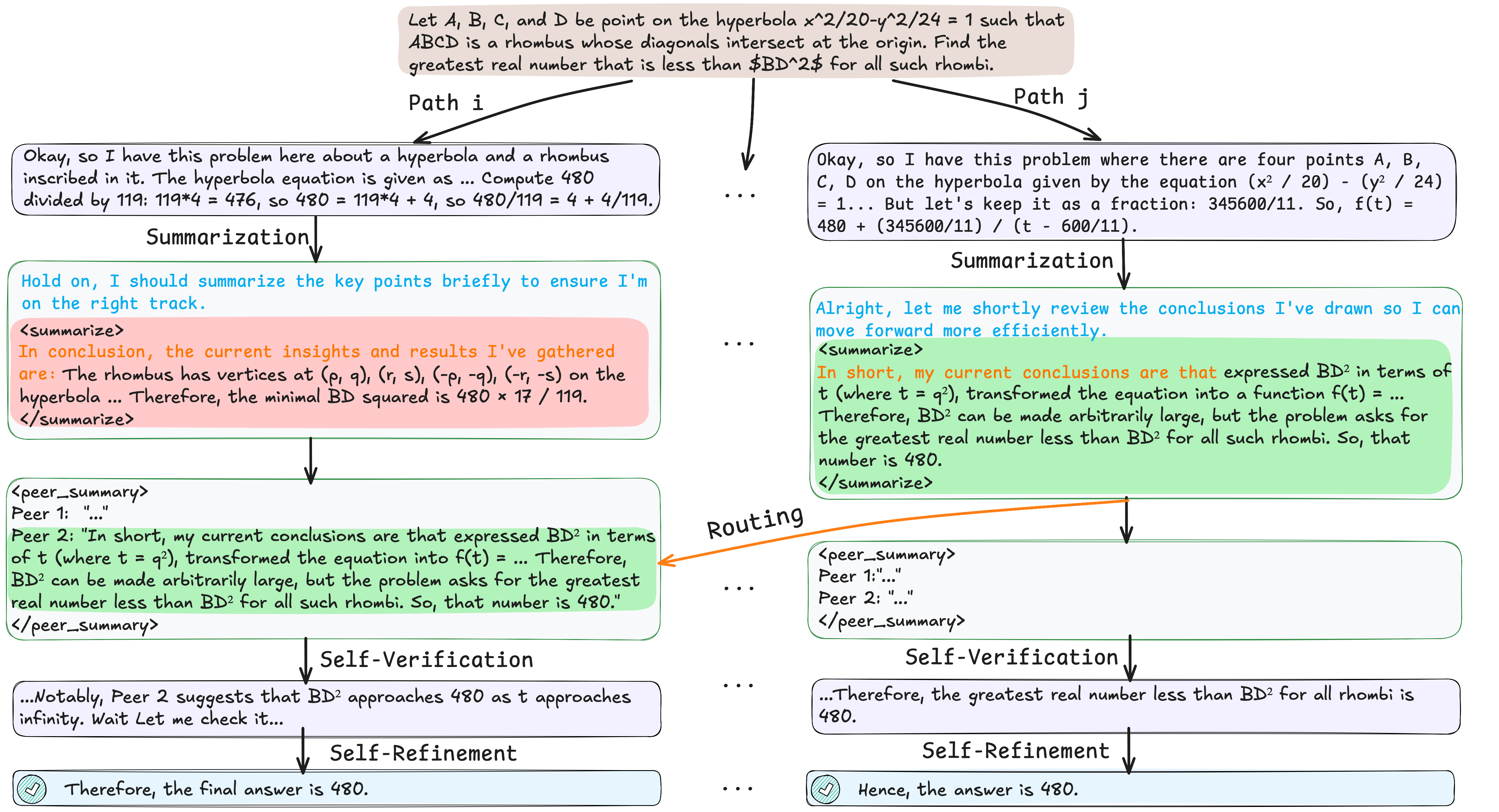}
    \caption{An example of how LeaP enables communication between path $i$ and $j$.
Text in \textcolor{darkred}{red} indicates the current path is incorrect.
Text in \textcolor{darkgreen}{green} shows a correct summary received from a peer.}
    \label{fig:communication_block} \vspace{-3mm}
\end{figure}

\noindent \textbf{Overview.}
As depicted in Figure~\ref{fig:main}b, LeaP distinguishes itself from traditional independent reasoning (Figure~\ref{fig:main}a) by strategically inserting LeaP blocks into the parallel inference process.
These blocks serve as dedicated modules for cross-path communication and collaborative refinement.
Each LeaP block orchestrates a two-stage process: (1) Summarization, where each reasoning path concisely articulates its current state, and (2) Routing, where these distilled insights are efficiently disseminated to peer paths.
An example of how peer insights can trigger self-correction is provided in Figure~\ref{fig:communication_block}.

\noindent \textbf{1. Summarization stage: generating shareable insights.}
Effective peer learning requires efficient insight sharing. The first stage in a LeaP block is Summarization. Each LRM path condenses its current approach, key insights, and intermediate results into a concise summary. This summary is strictly limited to 256 tokens to maintain token efficiency. A prompt, comprising a dynamically selected \textcolor{cyan}{summary trigger} and a \textcolor{orange}{summary template}, directs this summarization. These elements are randomly chosen from predefined lists (see Appendix~\ref{app:prompts}), promoting variability in summary expression while ensuring core information capture.
Once these concise summaries are generated by each path, the next step is to distribute them among peers. This is handled by the Routing stage.

\noindent \textbf{2. Routing stage: disseminating peer insights.}
Although summaries provide condensed information, exposing each path to all peer summaries is overwhelming and token-inefficient, especially with numerous parallel paths. The Routing stage mitigates this by selecting which peer summaries a given path receives. For $N$ reasoning paths and their summaries $\{s_1, s_2, \dots, s_N\}$, a routing function $\mathcal{R}$ selects for each path $i$ a subset of peer summaries $\mathcal{C}_i \subset \{ s_j \mid j \neq i \}$. We explore three routing mechanisms, each employing a different heuristic to foster effective collaboration:

\begin{itemize}
    \item \textbf{Dispersed routing:}
Grounded in the \textbf{intuition} that diverse insights are crucial for breaking out of erroneous reasoning patterns and discovering novel solutions, Dispersed Routing prioritizes summaries that are least similar to the receiving path's own summary:
\begin{eqnarray} \small
\mathcal{C}_i = 
\text{Bottom-}k
\left(
\left(
\texttt{similarity}(s_i, s_j) \mid j \neq i
\right)
\right)
\end{eqnarray}

    \item \textbf{Clustered routing:}
Conversely, the \textbf{assumption} here is that paths with similar reasoning are likely converging on a viable solution trajectory. Clustered Routing selects the top-$k$ most similar summaries, facilitating collaboration among paths that are already aligned:
\begin{eqnarray} \small
\mathcal{C}_i = \text{Top-}k\left(\left\{ \texttt{similarity}(s_i, s_j) \mid j \neq i \right\}\right)
\end{eqnarray}

    \item \textbf{Hybrid routing}
To achieve a synthesis of the above strategies, Hybrid Routing seeks to balance the benefits of collaborative reinforcement with the injection of diverse perspectives. It selects $\frac{k}{2}$ most similar summaries (exploitation) and $\frac{k}{2}$ most dissimilar summaries (exploration):
\begin{eqnarray} \small
\mathcal{C}_i = \mathcal{C}_i^{\text{sim}} \cup \mathcal{C}_i^{\text{dis}}, \quad \text{where } |\mathcal{C}_i^{\text{sim}}| = |\mathcal{C}_i^{\text{dis}}| = \frac{k}{2}
\end{eqnarray}
\end{itemize}

To quantify the similarity between summaries $s_i$ and $s_j$ for these routing mechanisms, we employ the normalized Levenshtein similarity~\cite{levenshtein1966binary}. The similarity score $\texttt{similarity}(s_i, s_j)$ is defined as:
\begin{eqnarray} \small
\texttt{similarity}(s_i, s_j) = 1 - \frac{D_{\text{lev}}(s_i, s_j)}{\max(|s_i|, |s_j|)}
\end{eqnarray}
where $D_{\text{lev}}(s_i, s_j)$ is the Levenshtein distance~\cite{levenshtein1966binary} between $s_i$ and $s_j$, and $|s_i|$ denotes the length of the string $s_i$. This score, ranging from $[0, 1]$, provides a normalized measure where higher values indicate greater textual similarity.

\vspace{-2mm}
\subsection{Can LeaP help in ``Prefix Dominance Trap''?}\label{sec:eval_prefix}
\vspace{-2mm}
\begin{wrapfigure}{R}{0.4\textwidth}
    \centering \vspace{-9mm}
    \includegraphics[width=\linewidth]{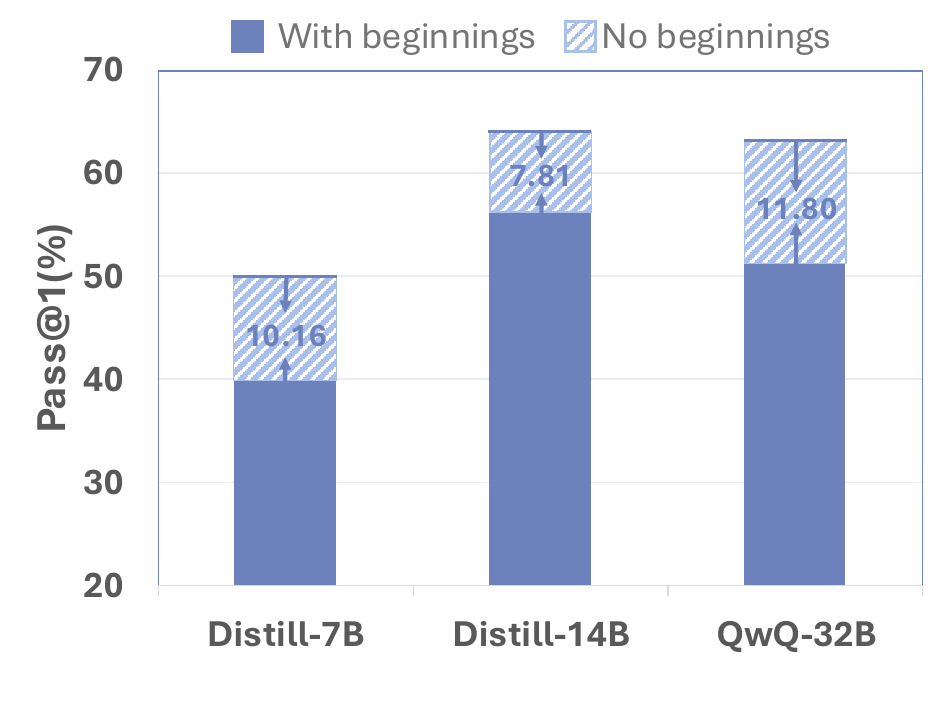}
    \caption{The results of starting with bad beginnings for models with LeaP.}
    \label{fig:motivation_ours} \vspace{-3mm}
\end{wrapfigure}
To verify if LeaP can help overcome this \emph{``Prefix Dominance Trap''}–the phenomenon where short, flawed beginnings significantly hinder recovery–we assess reasoning models with LeaP under the setting in Section~\ref{sec:motivation}.
As shown in Figure~\ref{fig:motivation_ours}, we present Pass@1 results of models with LeaP on the AIME 2024 subset from Section~\ref{sec:motivation}.
LeaP consistently reduces the performance gap across all three model sizes.
For example, on DeepSeek-Distill-Qwen-14B, LeaP narrows the gap from 19.88 to 7.81.
Appendix~\ref{app:case_study_bad_begin} provides case studies comparing the baseline with LeaP.
We also conduct experiments starting from good beginnings, where truncated initial tokens likely lead to correct answers.
As presented in Appendix~\ref{app:good_beginning}, when the beginning is good, our method also performs better than standard independent reasoning.
These results on both bad and good beginnings demonstrate that \textbf{LRMs leverage strong self-verification to assess expanded reasoning paths from peers}, improving LRM self-correction.

\vspace{-2mm}
\section{Comprehensive Evaluation of LeaP}\label{sec:eval}\vspace{-2mm}
\label{sec:eval_reasoning}

To assess LeaP comprehensively, we evaluate the LeaP on four reasoning benchmarks. 
Section~\ref{sec:eval_setup} describes the evaluation setup and implementation details.
In Section~\ref{sec:eval_results}, we report results showing the effectiveness of LeaP across four benchmarks.
\vspace{-2mm}
\subsection{Evaluation Setup}\label{sec:eval_setup}\vspace{-2mm}

\noindent \textbf{Benchmarks and metrics.}
We evaluate our method on challenging benchmarks.
Specifically, we use competition-level mathematics datasets: AIME 2024~\cite{aime2024}, AIME 2025~\cite{aime2025}, and the AIMO 2025~\cite{ai-mathematical-olympiad-progress-prize-2} reference set.
AIME 2024~\cite{aime2024} and AIME 2025~\cite{aime2025} each contain 30 problems; AIMO 2025~\cite{ai-mathematical-olympiad-progress-prize-2} contains 10 problems.
Additionally, we test on the GPQA~\cite{rein2024gpqa} Diamond subset (198 questions), the most difficult GPQA~\cite{rein2024gpqa} portion, requiring PhD-level domain-specific knowledge in physics, chemistry, and biology.
For evaluation, we adopt Pass@1 and Cons@N~\cite{guo2025deepseek}.
We generate $N$ responses per question using a non-zero temperature.
Pass@1 is calculated as:
$ \text{Pass@1} = \frac{1}{N} \sum_{i=1}^{N} \mathbb{I}_i $,
where $\mathbb{I}_i$ denotes $i$-th response correctness.
We also report consensus results by voting from $N$ responses, termed Cons@N.

\noindent \textbf{Models and baselines.} We select LRMs with various sizes from 7B to 32B, including the 7B and 14B versions of DeepSeek-R1-Distill-Qwen~\cite{guo2025deepseek} and QwQ-32B~\cite{qwq32b}.
We run these models using the vllm framework~\cite{kwon2023efficient}.
We use independent reasoning as the baseline for Pass@1, and adopt Majority Voting~\cite{wang2022self} and Shortest Majority Voting~\cite{zeng2025revisiting} as baseline methods for Cons@N.

\noindent \textbf{Hyperparameters.} Following the settings reported in QwQ-32B~\cite{qwq32b}, we use a sampling temperature of 0.6, a Top-p value of 0.95, and a Top-k value of 40 to generate $N$ responses. Typically, $N$ is set to 8 for GPQA~\cite{rein2024gpqa} tasks and 32 for math tasks~\cite{aime2024, aime2025, ai-mathematical-olympiad-progress-prize-2}, depending on the size of the benchmark. 
For the maximum tokens, we use 16,384 for the 7B and 14B models, and 32,768 for the 32B model.

\vspace{-2mm}
\subsection{Results}\label{sec:eval_results}\vspace{-2mm}

The results are presented in Table~\ref{tab:main_Pass@1}. Overall, LRMs with LeaP significantly outperform the baselines. For example, in terms of average performance across four benchmarks, our method using DeepSeek-R1-Distill-Qwen-7B with top-4 \textit{Dispersed} routing exceeds the baseline by 6.49. 
Similarly, on DeepSeek-R1-Distill-Qwen-14B, it surpasses the baseline by 6.08. \textbf{Particularly, QwQ-32B with top-4 \textit{Dispersed} routing even beats R1-671B in all three math datasets.}

We also observe performance improvements when transferring from Top-$2$ to Top-$4$ peer summaries. This trend suggests that as the number of top-$k$ peer summaries increases, the LRM benefits thinking paths from a greater number of peers.
Among three routing strategies, \textit{Dispersed}, and \textit{Hybrid} clearly outperform the \textit{Clustered} approach.
This is expected, as receiving similar summaries from peers limits the overall diversity of trajectories for self-verification.
In contrast, the diverse or complementary perspectives provided by \textit{Dispersed} and \textit{Hybrid} routing introduce a broader range of reasoning paths.
To provide a comprehensive evaluation, we also report the Cons@N results in Appendix~\ref{app:cons@N}

\vspace{-2mm}
\begin{table}[H]
    \centering
    \small
    \caption{We evaluate the baseline and LeaP on AIME 2024~\cite{aime2024}, AIME 2025~\cite{aime2025}, AIMO 2025~\cite{ai-mathematical-olympiad-progress-prize-2}, and GPQA Diamond~\cite{rein2024gpqa}.
    Our method significantly outperforms the baseline.
    We also report DeepSeek-R1-671B results to demonstrate our method's advantages.
    Since DeepSeek-R1-671B lacks an official AIMO 2025 score, we evaluate it on the official website, repeating inference four times.
    R1-7B and R1-14B represent the 7B and 14B versions of DeepSeek-R1-Distill-Qwen, respectively.
}
    \begin{tabular}{lc|ccccccc|c}
    \toprule
    \multirow{2}{*}{\textbf{Benchmarks}} &\multirow{2}{*}{\textbf{Models}} & \multirow{2}{*}{\textbf{Baseline}} &\multicolumn{2}{c}{\textbf{Clustered}}& \multicolumn{2}{c}{\textbf{Hybrid}} & \multicolumn{2}{c|}{\textbf{Dispersed}} & \multirow{2}{*}{\textbf{R1-671B}}\\
    
    \cline{4-9}
    & &  & \textbf{Top-2} & \textbf{Top-4} & \textbf{Top-2} & \textbf{Top-4}& \textbf{Top-2} & \textbf{Top-4} & \\
    
    \specialrule{0.5pt}{2pt}{2pt}
    \multirow{3}{*}{AIME 2024} 
    & R1-7B   & 
    \cellcolor{mygreen1}51.35 & 
    \cellcolor{mygreen2}56.15 &
    \cellcolor{mygreen3}\textcolor{white}{59.27} &
    \cellcolor{mygreen3}\textcolor{white}{59.17} &
    \cellcolor{mygreen4}\textcolor{white}{61.67} &
    \cellcolor{mygreen3}\textcolor{white}{60.31} &
    \cellcolor{mygreen3}\textcolor{white}{60.52} &
    \multirow{3}{*}{79.8} 
    \\
    & R1-14B  & 
    \cellcolor{mygreen1}64.47 & 
    \cellcolor{mygreen2}72.08&
    \cellcolor{mygreen2}{71.04}  &
    \cellcolor{mygreen2}71.77 &
    \cellcolor{mygreen3}\textcolor{white}{74.48} &
    
    \cellcolor{mygreen2}71.15 & 
    \cellcolor{mygreen4}\textcolor{white}{77.29}  &

    \\
    & QwQ-32B & \cellcolor{mygreen2}79.69 & 
    \cellcolor{mygreen1}78.96&
    \cellcolor{mygreen3}\textcolor{white}{81.56} &

    \cellcolor{mygreen3}\textcolor{white}{81.88} &
    \cellcolor{mygreen3}\textcolor{white}{81.67}&
    \cellcolor{mygreen3}\textcolor{white}{81.56} &
    \cellcolor{mygreen4}\textcolor{white}{85.83} 
    \\

    \specialrule{0.5pt}{2pt}{2pt}
    \multirow{3}{*}{AIMO 2025} 
    & R1-7B   & 
    \cellcolor{mygreen1}37.50  & 
    \cellcolor{mygreen1}37.81  &
    \cellcolor{mygreen2}{39.06}  &
    \cellcolor{mygreen3}\textcolor{white}{41.88}  &
    \cellcolor{mygreen3}\textcolor{white}{40.31}  &
    
    \cellcolor{mygreen2}39.06  & 
    \cellcolor{mygreen4}\textcolor{white}{45.00}  &

    \multirow{3}{*}{65.0} 
    \\
    & R1-14B  & \cellcolor{mygreen1}46.87 &
    \cellcolor{mygreen3}\textcolor{white}{51.50}&
    \cellcolor{mygreen4}\textcolor{white}{55.31}&

    \cellcolor{mygreen3}\textcolor{white}{51.88} & 
    \cellcolor{mygreen3}\textcolor{white}{52.19}&
    \cellcolor{mygreen2}49.38 &
    \cellcolor{mygreen3}\textcolor{white}{51.25} 
    \\
    & QwQ-32B &
    \cellcolor{mygreen1}63.75 &
    \cellcolor{mygreen1}63.75&
    \cellcolor{mygreen3}\textcolor{white}{65.63} &

    \cellcolor{mygreen2}64.06  &
    \cellcolor{mygreen2}64.69&
    \cellcolor{mygreen4}\textcolor{white}{67.19} &
    \cellcolor{mygreen4}\textcolor{white}{67.19} &\\
    
    \specialrule{0.5pt}{2pt}{2pt}
    \multirow{3}{*}{AIME 2025} 
    & R1-7B    & 
    \cellcolor{mygreen1}37.81   &
    \cellcolor{mygreen1}36.98&
    \cellcolor{mygreen1}{37.50}&
    \cellcolor{mygreen1}37.19  &
    \cellcolor{mygreen3}\textcolor{white}{39.27}&
    \cellcolor{mygreen4}\textcolor{white}{40.93}  &
    \cellcolor{mygreen2}{38.44}  &

    \multirow{3}{*}{70.0} 
    \\
    & R1-14B   & \cellcolor{mygreen1}48.64   &
    \cellcolor{mygreen2}50.42&
    \cellcolor{mygreen2}{51.88}&

    \cellcolor{mygreen4}\textcolor{white}{54.38}  &
    \cellcolor{mygreen4}\textcolor{white}{54.38}&
    \cellcolor{mygreen2}50.31 &
    \cellcolor{mygreen4}\textcolor{white}{54.17}  &
    \\
    & QwQ-32B  & \cellcolor{mygreen1}68.13   &
    \cellcolor{mygreen2}68.85 &
    \cellcolor{mygreen3}\textcolor{white}{71.35} &

    \cellcolor{mygreen3}\textcolor{white}{71.04}  &
    \cellcolor{mygreen4}\textcolor{white}{72.50}&
    \cellcolor{mygreen3}\textcolor{white}{70.83} &
    \cellcolor{mygreen3}\textcolor{white}{71.67}  &
    \\
    
    \specialrule{0.5pt}{2pt}{2pt}
    \multirow{2}{*}{GPQA} 
    & R1-7B   & \cellcolor{mygreen1}46.91 &
    \cellcolor{mygreen2}52.97&
    \cellcolor{mygreen2}52.65&
    \cellcolor{mygreen2}51.83 &
    \cellcolor{mygreen3}\textcolor{white}{53.47}&
    
    \cellcolor{mygreen3}\textcolor{white}{53.28} &
    \cellcolor{mygreen4}\textcolor{white}{55.56} &

    \multirow{3}{*}{71.5} \\

    \multirow{2}{*}{Diamond} & R1-14B  &
    \cellcolor{mygreen1}53.47 &
    \cellcolor{mygreen2}54.80&
    \cellcolor{mygreen4}\textcolor{white}{58.33}&

    \cellcolor{mygreen2}54.42 &
    \cellcolor{mygreen4}\textcolor{white}{57.89}&
    \cellcolor{mygreen3}\textcolor{white}{55.68} &
    \cellcolor{mygreen2}55.05 
    \\
    
    & QwQ-32B &
    \cellcolor{mygreen1}58.00 &
    \cellcolor{mygreen3}\textcolor{white}{65.03}&
    \cellcolor{mygreen3}\textcolor{white}{65.28}&

    \cellcolor{mygreen2}61.87 &
    \cellcolor{mygreen3}\textcolor{white}{65.21}&
    \cellcolor{mygreen4}\textcolor{white}{66.16} &
    \cellcolor{mygreen2}63.32
    \\

    \specialrule{0.5pt}{2pt}{2pt}
    \multirow{3}{*}{Avg.} 
    & R1-7B  &
    \cellcolor{mygreen1}43.39&
    \cellcolor{mygreen2}45.98& 
    \cellcolor{mygreen3}\textcolor{white}{47.12}&
    \cellcolor{mygreen3}\textcolor{white}{47.52}&	
    \cellcolor{mygreen4}\textcolor{white}{48.68}&
    \cellcolor{mygreen4}\textcolor{white}{48.40}&	
    \cellcolor{mygreen4}\textcolor{white}{49.88}&

    \multirow{3}{*}{71.58} 
    \\

    & R1-14B &     
    \cellcolor{mygreen1}53.36&
    \cellcolor{mygreen2}{57.20}&
    \cellcolor{mygreen4}\textcolor{white}{59.14}&

    \cellcolor{mygreen3}\textcolor{white}{58.11}&	
    \cellcolor{mygreen4}\textcolor{white}{59.74}&
    \cellcolor{mygreen2}56.63&	
    \cellcolor{mygreen4}\textcolor{white}{59.44}&\\

    & QwQ-32B & 
    \cellcolor{mygreen1}67.39&
    \cellcolor{mygreen2}69.15&	
    \cellcolor{mygreen3}\textcolor{white}{70.96}&

    \cellcolor{mygreen2}69.71&	
    \cellcolor{mygreen3}\textcolor{white}{71.02}&
        \cellcolor{mygreen3}\textcolor{white}{71.44}&	
    \cellcolor{mygreen4}\textcolor{white}{72.00}&\\
   
    \bottomrule
    \end{tabular}
    \label{tab:main_Pass@1}
\end{table}
\vspace{-2mm}

\begin{figure}[H]
  \centering \vspace{-3mm}
  \begin{subfigure}[b]{0.49\textwidth}
        \centering
        \includegraphics[width=\linewidth]{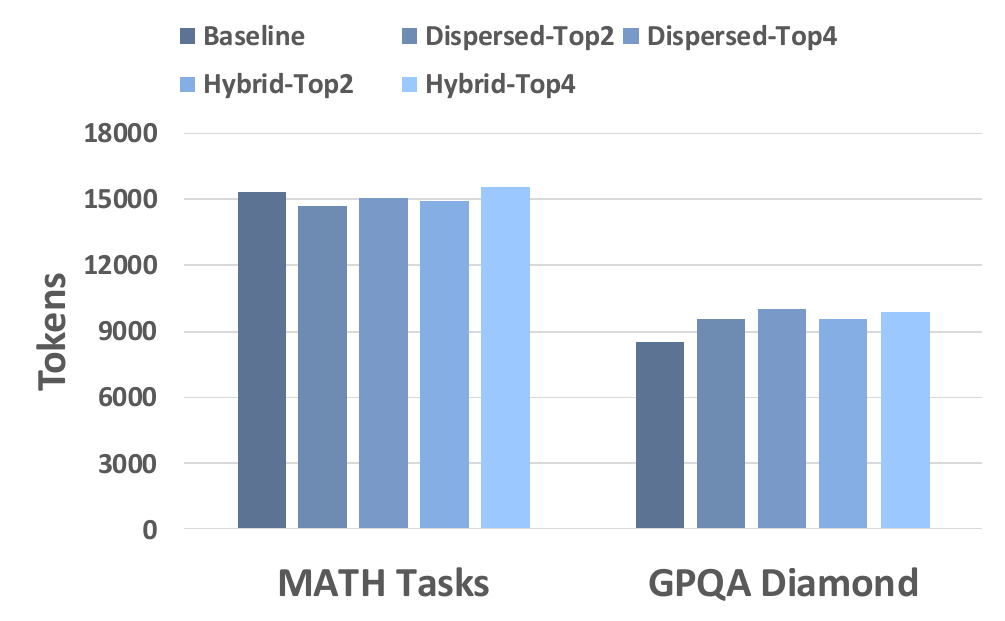}
        \caption{Number of Tokens}
        \label{fig:Tokens_32B}
    \end{subfigure}
  \begin{subfigure}[b]{0.49\textwidth}
        \centering
        \includegraphics[width=\linewidth]{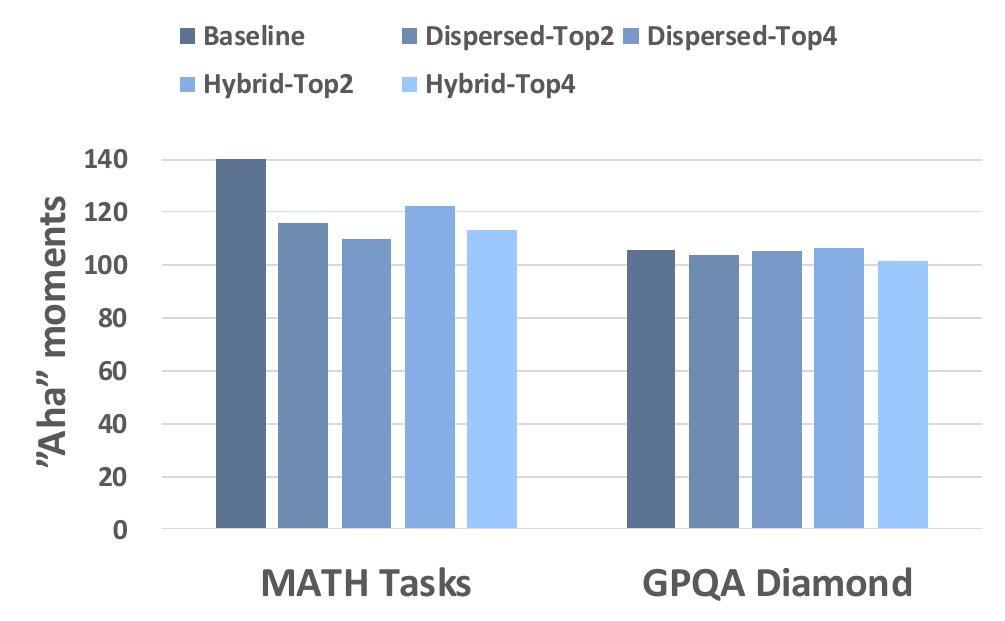}
        \caption{Number of ``Aha'' moments}
        \label{fig:Aha_32B}
    \end{subfigure}
    \caption{We illustrate the average number of tokens and ``Aha'' moments on QwQ-32B. Our method produces a comparable number of tokens to the baseline, while yielding fewer ``Aha'' moments.}
    \label{fig:main_eficiency} \vspace{-6mm}
\end{figure}

\noindent \textbf{Efficiency on generation token.}
We analyze LeaP's inference efficiency based on the total number of tokens.
Results for QwQ-32B~\cite{qwq32b} are presented in Figure~\ref{fig:main_eficiency}a; Appendix~\ref{app:efficiency} contains additional results.
LeaP does not generate significantly more tokens than the baseline; in fact, its average response length is sometimes shorter.
We further analyze ``aha'' moments using keywords from previous studies~\cite{zeng2025simplerl, li2025start, Slow_Thinking_with_LLMs_3} in Figure~\ref{fig:main_eficiency}b.
LeaP exhibits 16.4\% fewer ``aha'' moments than the baseline across three math benchmarks.
The number of ``aha'' moments for Top-$4$ settings is consistently lower than for Top-$2$ settings.
These findings indicate that receiving peer opinions and results reduces the model's need for reflection.

\vspace{-2mm}
\section{Empowering LeaP with Further Training Adaptation: LeaP-T}\vspace{-2mm}
By analyzing the responses in Section~\ref{sec:eval_reasoning}, we observe that reasoning models with small size, such as DeepSeek-R1-Distill-Qwen-7B, sometimes fail to summarize and reflect on peers' summaries effectively. The case studies can refer to Appendix~\ref{app:case_sutdy_failure}.
To this end, we introduce \textbf{LeaP-T} model series, where we attempt to alleviate this problem through supervised fine-tuning.

\vspace{-2mm}
\subsection{Experimental Setup}\label{sec:leap_t_setup}\vspace{-2mm}
We use approximately 1,000 AIME problems from 1984 to 2023~\cite{aime2024, aime2025} as source data.
We synthesize responses by applying LeaP to DeepSeek-R1-Distill-Qwen-32B~\cite{guo2025deepseek} and filter suitable responses using a rule-based selection mechanism.
We use supervised finetuning to train our LeaP-T models, starting from the 1.5B, 7B, and 14B versions of DeepSeek-R1-Distill-Qwen~\cite{guo2025deepseek}.
Specifically, a response is selected if its final answer is correct and the length of all summaries is less than $256$ tokens.
We also generate SFT data without LeaP to train baseline models.
Training hyperparameters are in Appendix~\ref{app:hyper_leapt}.
We use the hyperparameters and metrics setting from Section~\ref{sec:eval_setup}.
For LeaP, we use the Top-$4$ \textit{Dispersed} routing setting.

\vspace{-2mm}
\subsection{Results}\label{sec:leap_t_results}\vspace{-2mm}

\begin{table}[H]
    \centering
    \small \vspace{-3mm}
    \caption{Evaluation of our LeaP-T from 1.5B to 14B on three math benchmarks.}
    \begin{tabular}{l|cccccccc}
    \toprule
    \multirow{2}{*}{\textbf{Models}}  & \multicolumn{2}{c}{\textbf{AIME 2024}} & \multicolumn{2}{c}{\textbf{AIME 2025}} & \multicolumn{2}{c}{\textbf{AIMO 2025}} & \multicolumn{2}{c}{\textbf{Avg.}} \\
    \cline{2-9}
    & \scriptsize Pass@1 & \scriptsize Cons@32 & \scriptsize Pass@1 & \scriptsize Cons@32 & \scriptsize Pass@1 & \scriptsize Cons@32 & \scriptsize Pass@1 & \scriptsize Cons@32 \\
    \specialrule{0.5pt}{2pt}{2pt}
    R1-1.5B & 32.00 & 50.00 & 24.69 & 30.00 & 14.00 & 30.00 & 23.56 & 36.67 \\
    + SFT & 31.04 & \textbf{56.67} & 23.23 & 36.67 & 15.31 & 30.00 & 23.19 & 41.11 \\
    \rowcolor{mygreen2} + LeaP & 34.90 & 46.67 & 26.46 & 30.00 & 15.63 & 30.00 & 25.66 & 35.56\\
    \rowcolor{mygreen2} LeaP-T-1.5B & \textbf{37.08} & \textbf{56.67} & \textbf{26.67} & \textbf{40.00} & \textbf{20.31} & \textbf{40.00} & \textbf{28.02} & \textbf{45.56}\\
    \specialrule{0.5pt}{2pt}{2pt}
    R1-7B & 51.35 & 73.33 & 37.81 & 50.00 & 37.50 & 50.00 & 42.22 & 57.78\\
    + SFT & 51.56 & \textbf{80.00} & 35.73 & 53.33 & 33.75 & 40.00 & 40.35 & 57.78\\
    \rowcolor{mygreen2} + LeaP & 60.52 & 76.67 & 38.44 & 53.33 & \textbf{45.00} & 50.00 & 47.99 & 60.00\\
    \rowcolor{mygreen2} LeaP-T-7B & \textbf{64.38} & \textbf{80.00} & \textbf{41.25} & \textbf{56.67} & 44.06 & \textbf{60.00} & \textbf{49.90} & \textbf{65.56} \\
    \specialrule{0.5pt}{2pt}{2pt}
    R1-14B & 64.47 & 80.00 & 48.64 & 60.00 & 46.87 & 60.00 & 53.33 & 66.67\\
    + SFT & 65.63 & \textbf{83.33} & 46.88 & 63.33 & 45.63 & 60.00 & 52.71 & 68.89\\
    \rowcolor{mygreen2} + LeaP & \textbf{77.29} & \textbf{83.33} & 54.17 & 60.00 & 51.25 & 60.00 & 60.90 & 67.78\\
    \rowcolor{mygreen2} LeaP-T-14B & 76.46 & \textbf{83.33} & \textbf{54.27} & \textbf{70.00} & \textbf{52.50} & 60.00 & \textbf{61.08} & \textbf{71.11}\\
    \bottomrule
    \end{tabular}
    \label{tab:LeaP_T_main}
\end{table}
\vspace{-2mm}

\begin{wrapfigure}{R}{0.4\textwidth}
    \centering \vspace{-6mm}
    \includegraphics[width=\linewidth]{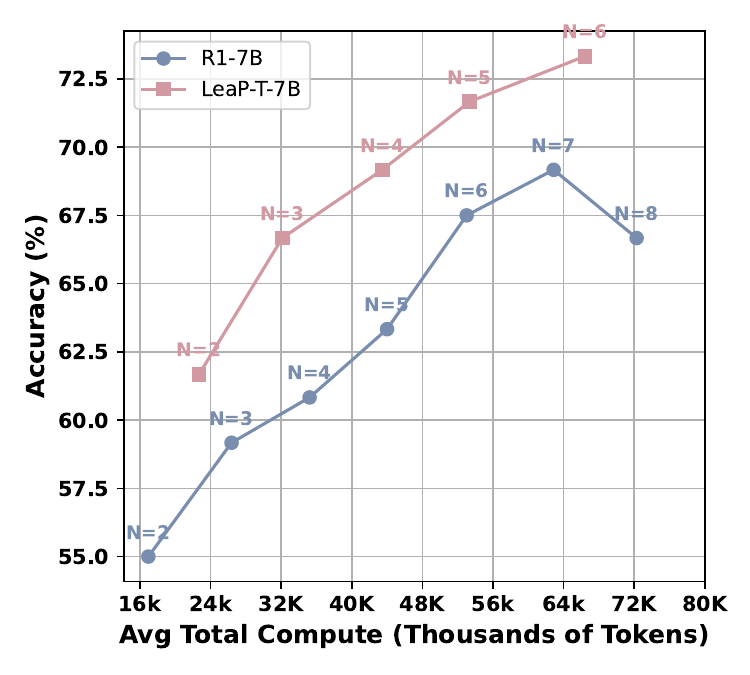}
    \caption{Accuracy vs. Total tokens on AIME 2024. LeaP-T-7B achieves higher accuracy compared to independent parallel reasoning under similar budgets.}
    \label{fig:scaling} \vspace{-3mm}
\end{wrapfigure}
Table~\ref{tab:LeaP_T_main} presents Pass@1 and Cons@32 results from DeepSeek-R1-Distill-Qwen-1.5B to DeepSeek-R1-Distill-Qwen-14B on three math benchmarks.
The results show consistent improvement across various model sizes.
For example, LeaP-T-1.5B outperforms DeepSeek-R1-Distill-Qwen-1.5B by 4.45 in average Pass@1 and 8.89 in average Cons@32.
The 7B and 14B models exhibit a similar trend.
Additionally, comparing DeepSeek-R1-Distill-Qwen models with LeaP to LeaP-T models, we observe that LeaP-T models demonstrate stronger performance.
To exclude the impact of distilling with a 32B model, we conduct distillation using SFT without LeaP.
Surprisingly, models distilled by DeepSeek-R1-Distill-Qwen-32B do not show improvements, and in some cases, show a marginal decrease in performance.
This is reasonable as the DeepSeek-R1-Distill-Qwen 1.5B, 7B, and 14B models are already well-trained by distilling from R1-671B.
Thus, distilling them with DeepSeek-R1-Distill-Qwen-32B does not yield additional benefits.
In other words, our training incorporates the \textit{Learning from Peers} paradigm, rather than simply transferring knowledge through distillation.

We also assess the efficient test-time scaling ability of LeaP-T by comparing it with the baseline under the 7B size on AIME 2024~\cite{aime2024}.
As shown in Figure~\ref{fig:scaling}, we increase inference tokens by parallel generating multiple responses.
For each point in Figure~\ref{fig:scaling}, we report average results from four repeated runs.
It is clear that our LeaP-T-7B consistently outperforms DeepSeek-R1-Distill-Qwen-7B.
This result demonstrates that our method scales more efficiently during test-time.
We also compare our LeaP-T with MoA~\cite{wang2024mixture}; details can be found in Appendix~\ref{app:moa}.

\vspace{-2mm}
\section{In-depth Analysis of LeaP}\vspace{-2mm}
To gain a deeper understanding of how and why LeaP works, we conduct a comprehensive analysis from three aspects: communication sensitivity~(Section~\ref{sec:analysis_sensitivity}), robustness~(Section~\ref{sec:robustness}), and human verification~(Section~\ref{sec:analysis_human}). This analysis not only guides the practical deployment of LeaP, but also sheds light on the inner workings of learning from peers.

\vspace{-2mm}
\subsection{Sensitivity Analysis of Communication}\label{sec:analysis_sensitivity}\vspace{-2mm}
To better deploy LeaP in practice, we investigate the sensitivity of communication on several factors, including granularity $T$ (Section~\ref{sec:analysis_granularity}), traffic (Section~\ref{sec:analysis_traffic}), evolution tendency of types (Section~\ref{sec:analysis_type}), and position (Section~\ref{sec:analysis_position}).

\vspace{-2mm}
\subsubsection{On Communication Granularity $T$}\label{sec:analysis_granularity}\vspace{-2mm}

\begin{figure}[H]
  \centering \vspace{-3mm}
  \begin{subfigure}[b]{0.49\textwidth}
        \centering
        \includegraphics[width=\linewidth]{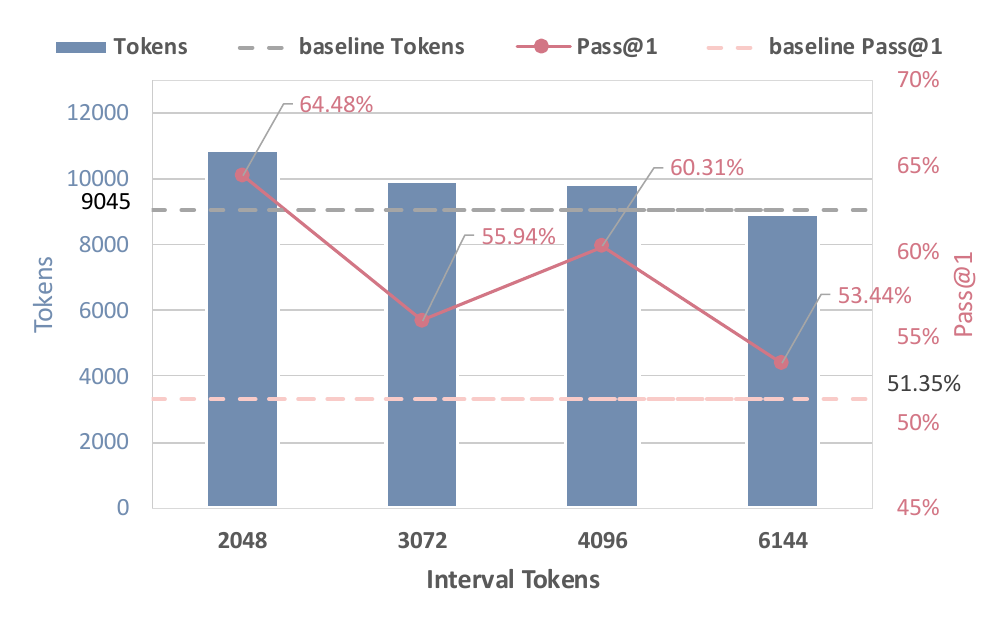}
        \caption{7B}
        \label{fig:Interval_Tokens_7B}
    \end{subfigure}
  \hspace{-2mm}
  \begin{subfigure}[b]{0.49\textwidth}
        \centering
        \includegraphics[width=\linewidth]{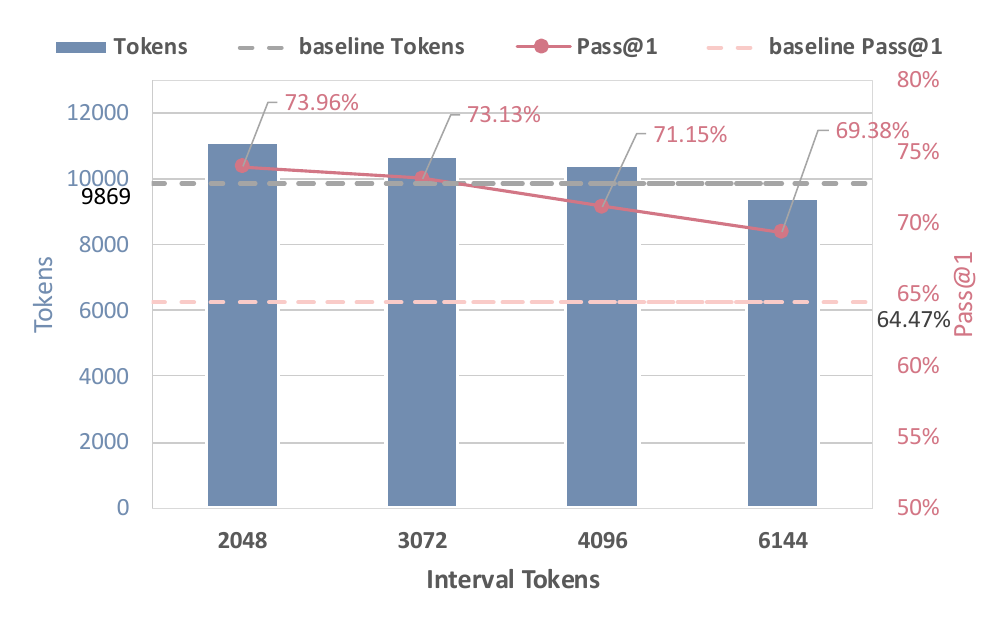}
        \caption{14B}
        \label{fig:Interval_Tokens_14B}
    \end{subfigure}
    \caption{Pass@1 and total tokens on AIME 2024 for 7B and 14B models with LeaP, evaluated across interval tokens of LeaP from 2048 to 6144.}
    \label{fig:ablation_freguency} \vspace{-3mm}
\end{figure}
To study the impact of communication granularity, we fix the top-$2$ \textit{Dispersed} routing strategy and vary the interval \( T \) between two consecutive LeaP blocks on the 7B and 14B versions of DeepSeek-R1-Distill-Qwen.
As shown in Figure~\ref{fig:ablation_freguency}, increasing the granularity of communication between reasoning paths slightly decreases performance on AIME 2024.
Specifically, the Pass@1 score drops from 64.48 to 53.44 for the 7B model, and from 73.96 to 69.38 for the 14B model.
Meanwhile, fewer tokens are consumed as communication becomes less sparse.
This is expected, since each LeaP block involves summarization and message routing, which costs additional tokens.
Overall, frequent communication (i.e., smaller \( T \)) slightly improves performance at the expense of more token usage, highlighting a trade-off between accuracy and efficiency.

\begin{wrapfigure}{R}{0.4\textwidth}
    \centering \vspace{-6mm}
    \includegraphics[width=\linewidth]{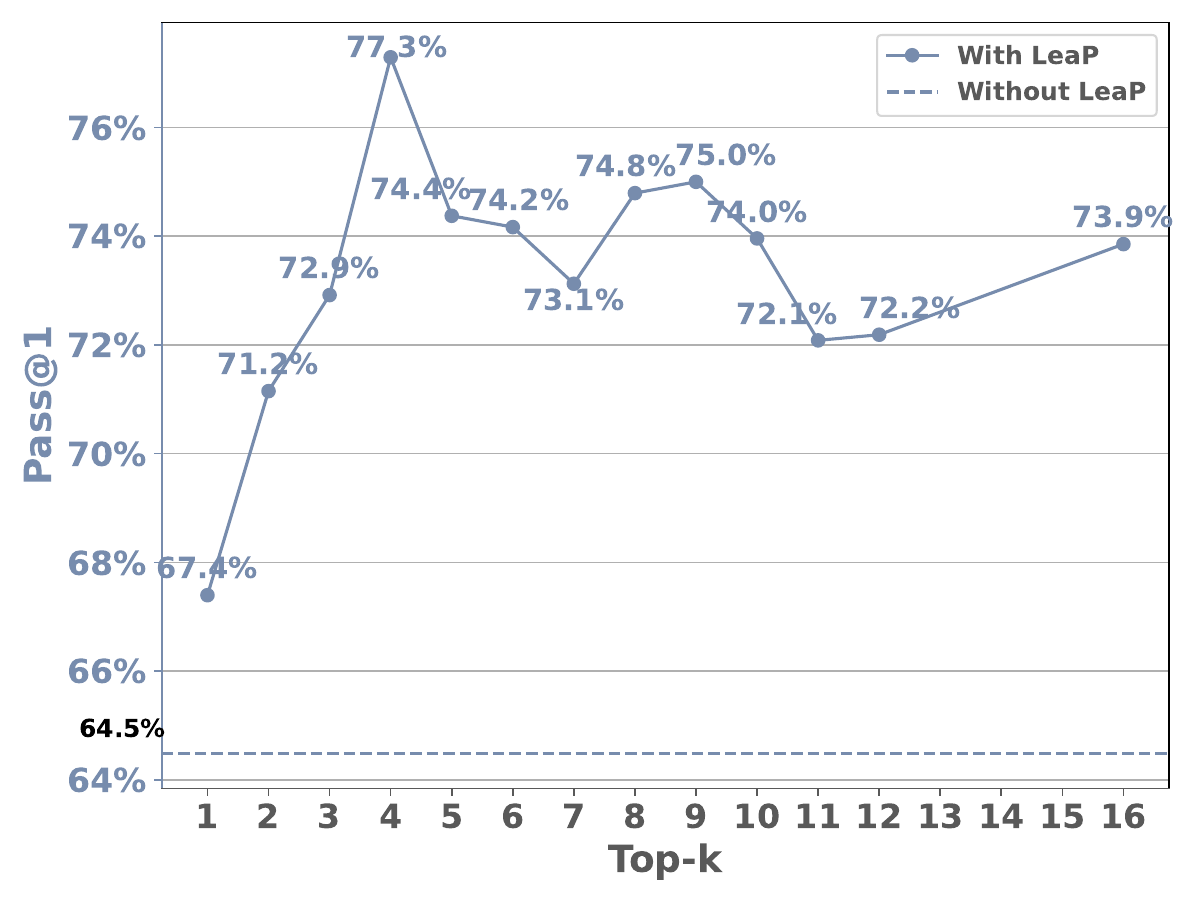}
    \caption{Performance of the 14B model with LeaP on AIME 2024, evaluated across the top-$k$ from $1$ to $16$.}
    \label{fig:top_k} \vspace{-3mm}
\end{wrapfigure}
\vspace{-2mm}
\subsubsection{On the Communication Traffic (Top-k)}\label{sec:analysis_traffic}
\vspace{-1mm}
For communication traffic, a natural assumption is that increasing it—i.e., routing more summaries from other reasoning paths—leads to better performance, as the model gains access to more diverse information.
However, our experiments reveal this intuition does not always hold.
To investigate this, we vary the number of routed summaries $k$ from 1 to 16 under the \textit{Dispersed} routing strategy, while fixing communication granularity $T$ to 4K tokens.
As shown in Figure~\ref{fig:top_k}, we evaluate DeepSeek-R1-Distill-Qwen-14B on AIME 2024 and observe a clear trend: the Pass@1 score rises sharply from Top-$1$ to Top-$4$, reaching its peak at $k=4$.
Surprisingly, performance then declines and fluctuates as $k$ increases beyond 4.
This finding suggests that larger communication traffic is not always better.
When $k$ is too small, the model lacks sufficient information, leading to limited perspectives.
On the other hand, a large $k$ introduces excessive summaries that may overwhelm the reasoning process, especially under constrained token budgets.
The resulting noise can impair coherence and hinder solution quality.
Overall, a trade-off exists between communication richness and cognitive overload, with Top-$4$ striking the most effective balance.

\begin{wrapfigure}{R}{0.4\textwidth}
    \centering \vspace{-1mm}
    \includegraphics[width=\linewidth]{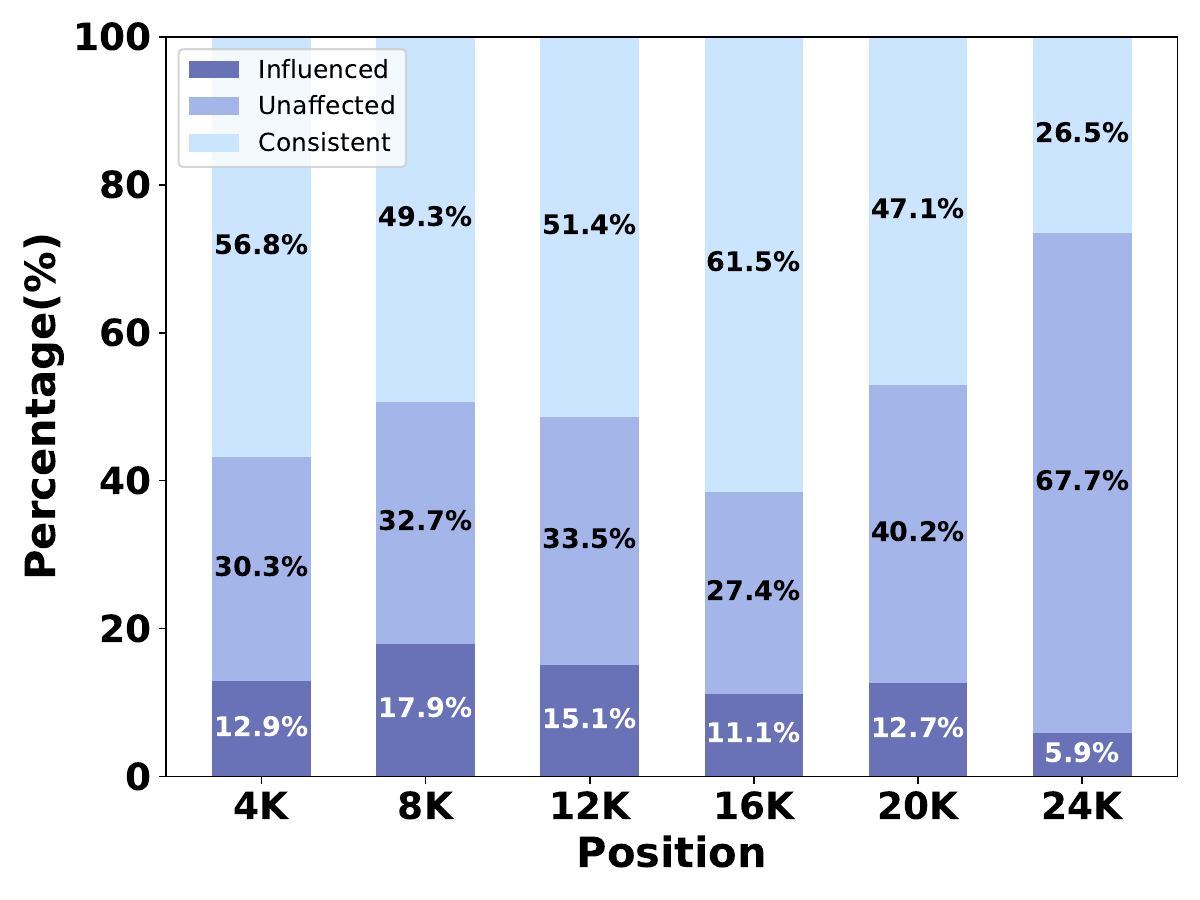}
    \caption{Communication types distribution of QwQ-32B on AIME 2024 at different positions, categorized into \textit{Consistent}, \textit{Unaffected}, and \textit{Influenced}.}
    \label{fig:analysis_position} \vspace{-4mm}
\end{wrapfigure}
\vspace{-1mm}
\subsubsection{On Evolution Tendency of Communication Types}\label{sec:analysis_type}
\vspace{-1mm}
Having established the optimal communication traffic (Top-k) as discussed previously, we further investigated the evolution of communication \textit{types} during reasoning to more comprehensively understand the dynamic characteristics of peer learning.
For this analysis, we fixed communication granularity at \( T = 4\text{K} \) tokens and used the top-$4$ \textit{Dispersed} routing strategy.
Communication outcomes were categorized as: \textit{Consistent} (paths align with peers before and after communication); \textit{Unaffected} (paths differ from peers and remain unchanged); or \textit{Influenced} (paths initially differ but adjust after receiving peer summaries).
GPT-4o~\cite{openai2024gpt4o} annotated these types, with human validation on 10\% of samples (approximately 120) confirming over 95\% model-human agreement.
Figure~\ref{fig:analysis_position} shows this type distribution for QwQ-32B on AIME 2024 (from 4K to 24K tokens).
\textit{Unaffected} cases increase significantly later, indicating reduced communication effectiveness as reasoning progresses.
Conversely, the \textit{Influenced} ratio, after rising from 0.13 (4K) to 0.18 (8K), steadily declines to 0.06 (24K).
This suggests communication is most impactful in early to mid-stage reasoning, with limited influence later.

\begin{wrapfigure}{R}{0.4\textwidth}
    \centering \vspace{-3mm}
    \includegraphics[width=\linewidth]{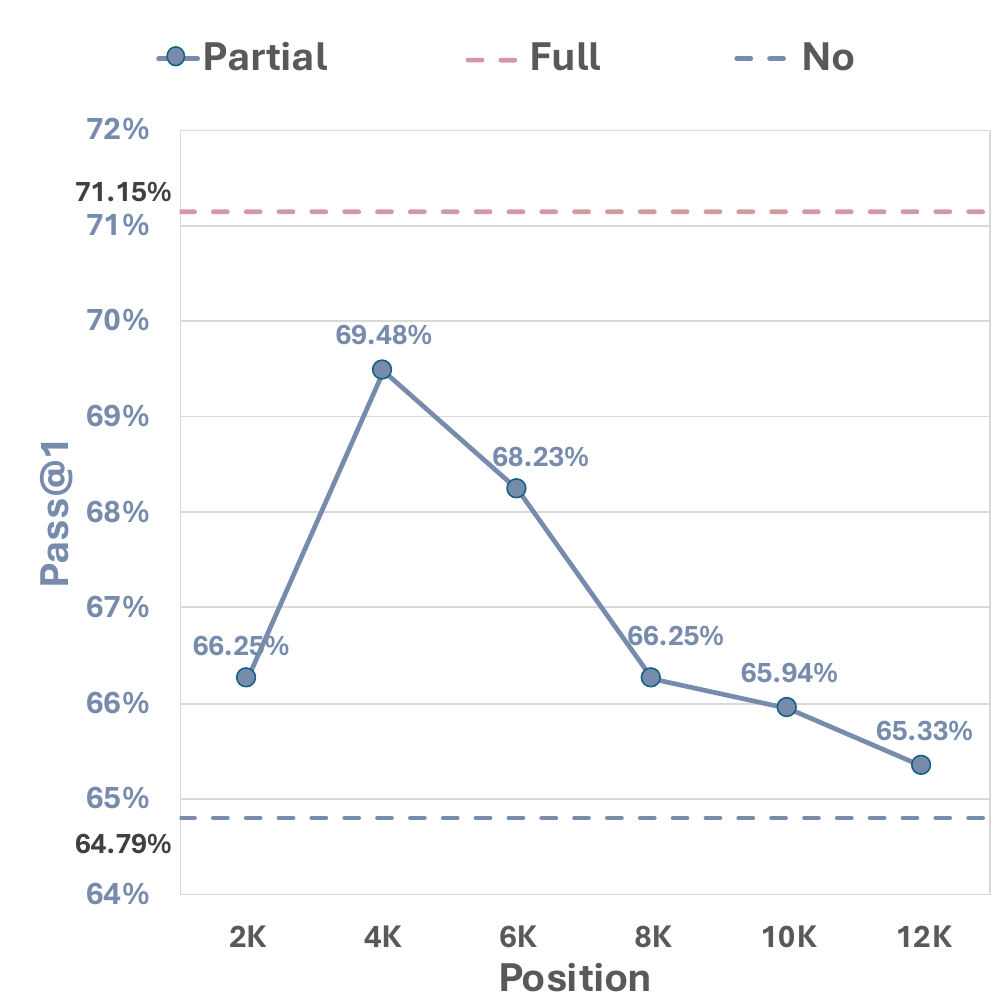}
    \caption{Performance of partial LeaP on the 14B model in inference, evaluated across the positions of LeaP block ranging from 2K to 12K tokens}
    \label{fig:Position_14B} \vspace{-3mm}
\end{wrapfigure}
\subsubsection{On the Communication Position (When only Communicating Once)}\label{sec:analysis_position}
\textbf{Setting on a new LeaP variant in single communication.}
To further analyze sensitivity to communicating at different positions, we introduce a simplified variant where communication occurs only once during the reasoning process.
This setting allows us to investigate the impact of communication position: whether it is more beneficial to communicate early, in the middle, or late.
In Figure~\ref{fig:Position_14B}, we report results of DeepSeek-R1-Distill-Qwen-14B~\cite{guo2025deepseek} on AIME 2024~\cite{aime2024} under this single communication setting.
We observe that performance improves when communication occurs early in the reasoning process, increasing from 66.25 at 2K tokens to a peak of 69.48 at 4K tokens.
However, beyond this point, performance declines, dropping to 65.33 by the end.
These results suggest that early-stage communication is more effective than late-stage interaction.
These findings are consistent with our previous conclusions from Section~\ref{sec:analysis_type}.
In particular, enabling a single LeaP block early in reasoning already yields substantial improvements over the baseline.
We refer to this variant as \textit{LeaP in Single Communication}.

\begin{table}[H]
    \centering \vspace{-3mm}
    \small
    \caption{The Pass@1 results of independent reasoning and LeaP in singler-generation on AIME 2024, AIME 2025, AIMO 2025 and GPQA Diamond for 14B model.}
    \begin{tabular}{l|ccc}
    \toprule
    \textbf{Benchmarks} & R1-14B & \textbf{$T=2$K} & \textbf{$T=4$K} \\
    \specialrule{0.5pt}{2pt}{2pt}
    AIME 2024 & 64.47 & 66.25 & \textbf{68.13}\\
    AIMO 2025 & 46.87 & 49.38 & \textbf{50.94}\\
    AIME 2025 & \textbf{48.64} & 45.83 & 47.29\\
    GPQA Diamond & 53.47 & 55.11 & \textbf{57.42} \\
    Avg. & 53.86 & 54.14 & \textbf{55.95} \\
    \bottomrule
    \end{tabular}
    \label{tab:serial_Pass@1}\vspace{-3mm}
\end{table}
\noindent \textbf{Results.}
We evaluate a simplified LeaP variant that retains only the first LeaP block at $T$ tokens and continues generation along a single path.
This mirrors how humans often work: gathering ideas early, then independently verifying them.
As shown in Table~\ref{tab:serial_Pass@1}, this single communication variant outperforms DeepSeek-R1-Distill-Qwen-14B.
For example, on AIME 2024, it improves Pass@1 from 64.47 to 68.13; on GPQA Diamond, from 53.47 to 57.42.
Even with just one block, it achieves a higher average Pass@1 (55.95 vs. 53.86), showing that early peer exposure effectively guides reasoning.
Gains are most notable in math and multi-hop tasks, where early external signals reduce error accumulation.
This variant also balances performance and efficiency, making it suitable for single-path scenarios.
We further explore different hyperparameters on AIME 2024 (Table~\ref{tab:14B_aime_serial_Pass@1}).
Increasing the interval $T$ from 2K to 4K improves accuracy (e.g., 67.92 to 69.48 at $k=8$), and using more peers (e.g., $k=8$ vs. $k=2$ at $T=4\text{K}$) also yields better results.

\begin{table}[H]
    \centering
    \small
    \caption{The Pass@1 results for varying $k$ values and position of LeaP block on AIME 2024 with the 14B model in single communication LeaP.}
    \begin{tabular}{l|cccccc}
    \toprule
    & {$k=2$} & {$k=4$} & {$k=6$} & {$k=8$} & {$k=16$ (Top-$2$)} & {$k=16$ (Top-$4$)} \\
    \specialrule{0.5pt}{2pt}{2pt}
    {$T=2$k} & 65.21 & 66.25 & 65.73 & 67.92 & 66.46 & 66.86 \\
    {$T=4$k} & 65.73 & 68.13 & 68.85 & 69.48 & 67.92 & 69.27 \\
    \bottomrule
    \end{tabular}
    \label{tab:14B_aime_serial_Pass@1}\vspace{-4mm}
\end{table}

\subsection{Post-hoc Robustness Analysis of LeaP}\label{sec:robustness}
To better understand the reliability of LeaP in practical scenarios, we conduct a post-hoc robustness analysis.
Specifically, we analyze whether the benefits of LeaP persist under two settings:
(1) Vary the error path in peers (Section~\ref{sec:robustness_error}), and (2) Vary the difficulty levels (Section~\ref{sec:robustness_difficulty}).

\subsubsection{Robustness on Error Tolerance}\label{sec:robustness_error}
\begin{wrapfigure}{R}{0.4\textwidth}
    \centering \vspace{-3mm}
    \includegraphics[width=\linewidth]{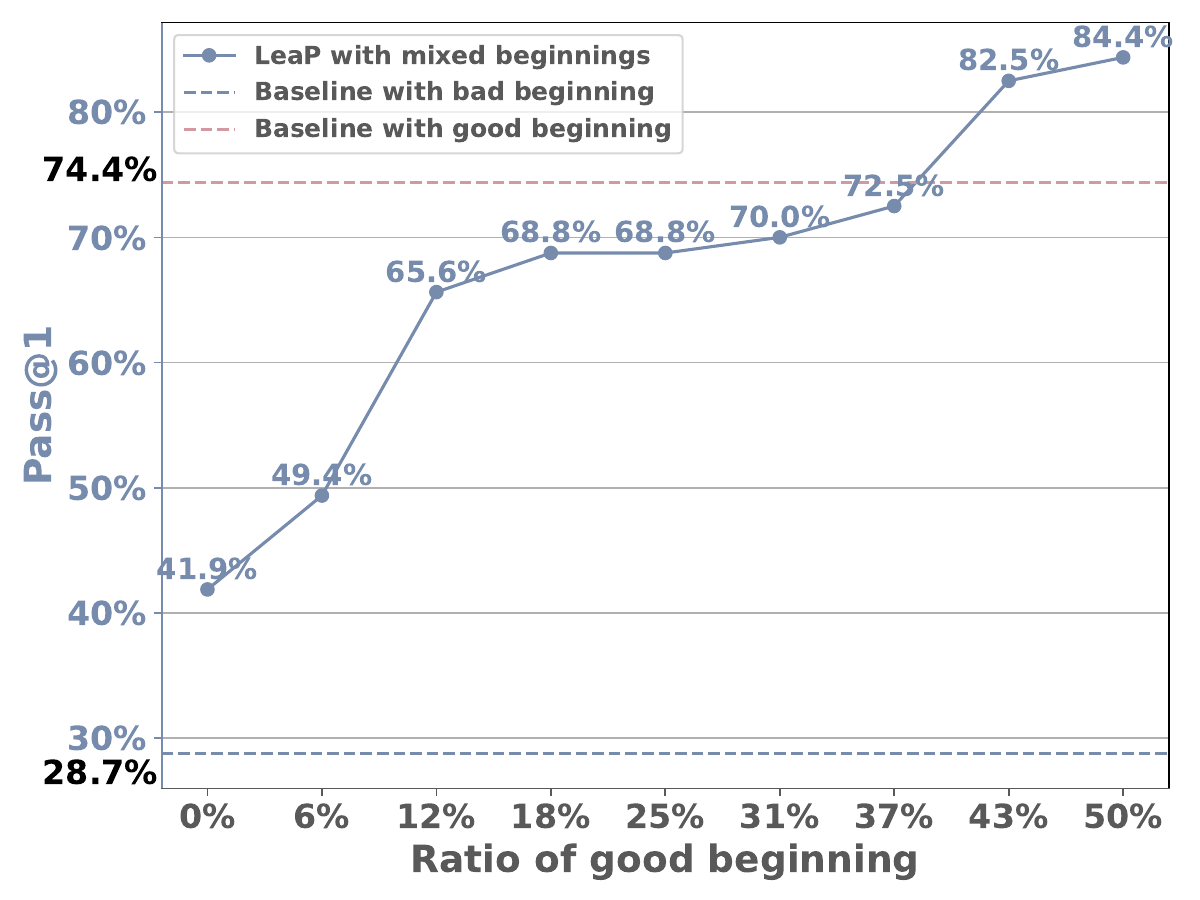}
    \caption{The Pass@1 under various ratios of good beginnings, where the length of the beginnings is 30\% of the average response length.}
    \label{fig:good_proportion} \vspace{-6mm}
\end{wrapfigure}
A common concern for LeaP is that low-quality paths may mislead others, especially when the majority of them are incorrect.
Intuitively, one might expect that effective collaboration in LeaP requires a high proportion of correct reasoning paths; otherwise, noise from incorrect paths could dominate interaction and degrade performance.

To test this hypothesis and assess LeaP's error tolerance, we design an experiment by varying the proportion of good beginnings.
Specifically, we select 10 questions with good and bad beginnings using the first 30\% tokens of responses for DeepSeek-R1-Distill-Qwen-14B by the method in Section~\ref{sec:motivation}.
For each question, we construct 16 mixed initial responses, varying the percentage of good beginnings from 0
As shown in Figure~\ref{fig:good_proportion}, the results reveal a surprising trend.
Across all configurations, the model with LeaP consistently outperforms the baseline, even when \textbf{none} of the beginnings are correct.
For instance, at 0\% good beginnings, LeaP achieves a Pass@1 score of 41.88, far exceeding the baseline score of 28.75.
Even more remarkably, at just 43\% good beginnings, LeaP surpasses the baseline's performance when \textbf{all} beginnings are good.
These results challenge the assumption that effective collaboration in LeaP requires mostly correct paths.
Instead, they demonstrate that LeaP is highly robust: peers can distill useful signals from noisy summaries.
This can be attributed to the strong self-verification ability of LRMs.

\subsubsection{Robustness at Various Difficulty Levels}\label{sec:robustness_difficulty}
Another concern is whether LeaP maintains robustness across various difficulty levels.
It is possible that LeaP performs well primarily because the benchmark contains a certain proportion of simple questions.
But can LeaP still provide benefits when problems become significantly more challenging?

\begin{figure}[H]
  \centering \vspace{-3mm}
  \begin{subfigure}[b]{0.49\textwidth}
        \centering
        \includegraphics[width=\linewidth]{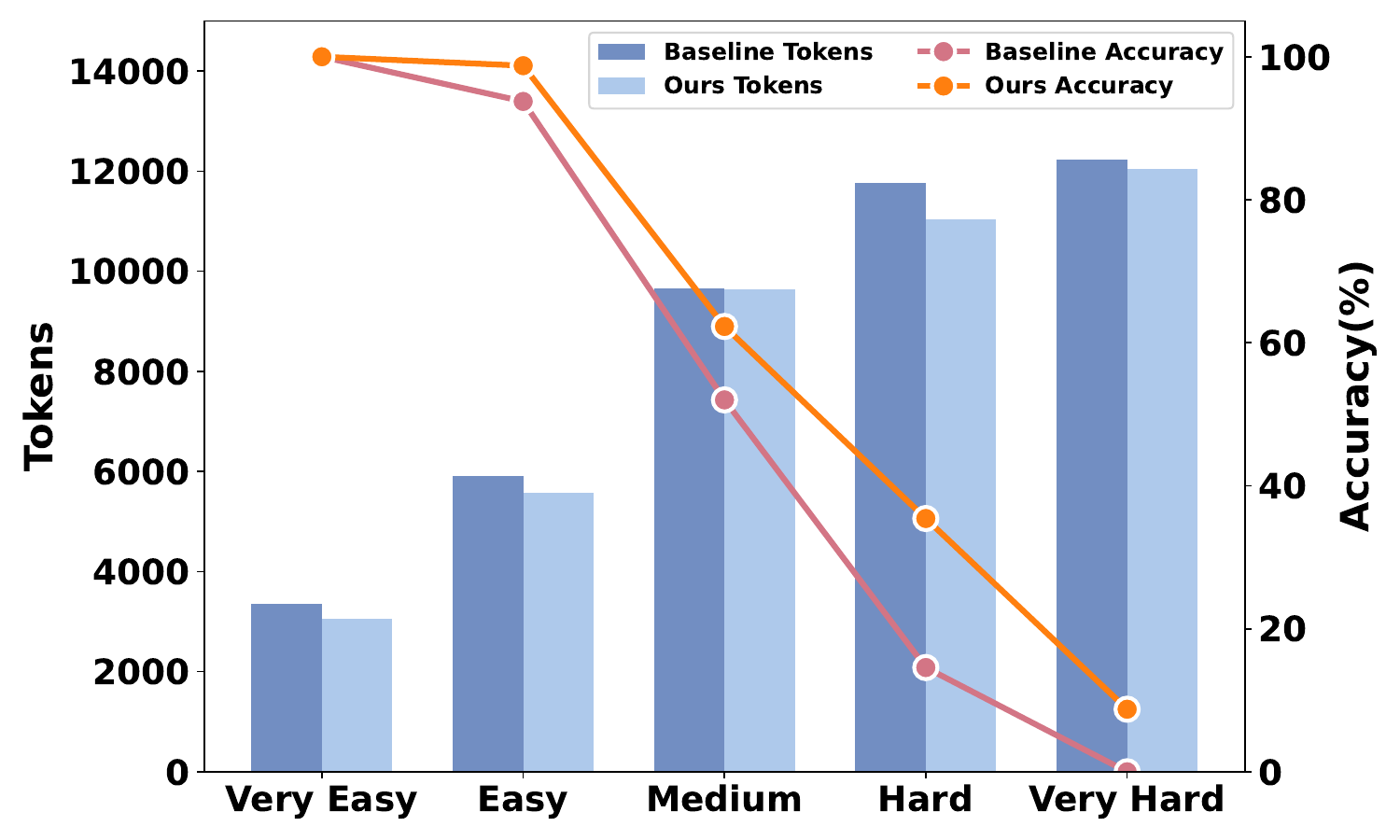}
        \caption{DeepSeek-R1-Distill-Qwen-7B}
        \label{fig:difficulty_leap_i_7B}
    \end{subfigure}
  \hspace{-2mm}
  \begin{subfigure}[b]{0.49\textwidth}
        \centering
        \includegraphics[width=\linewidth]{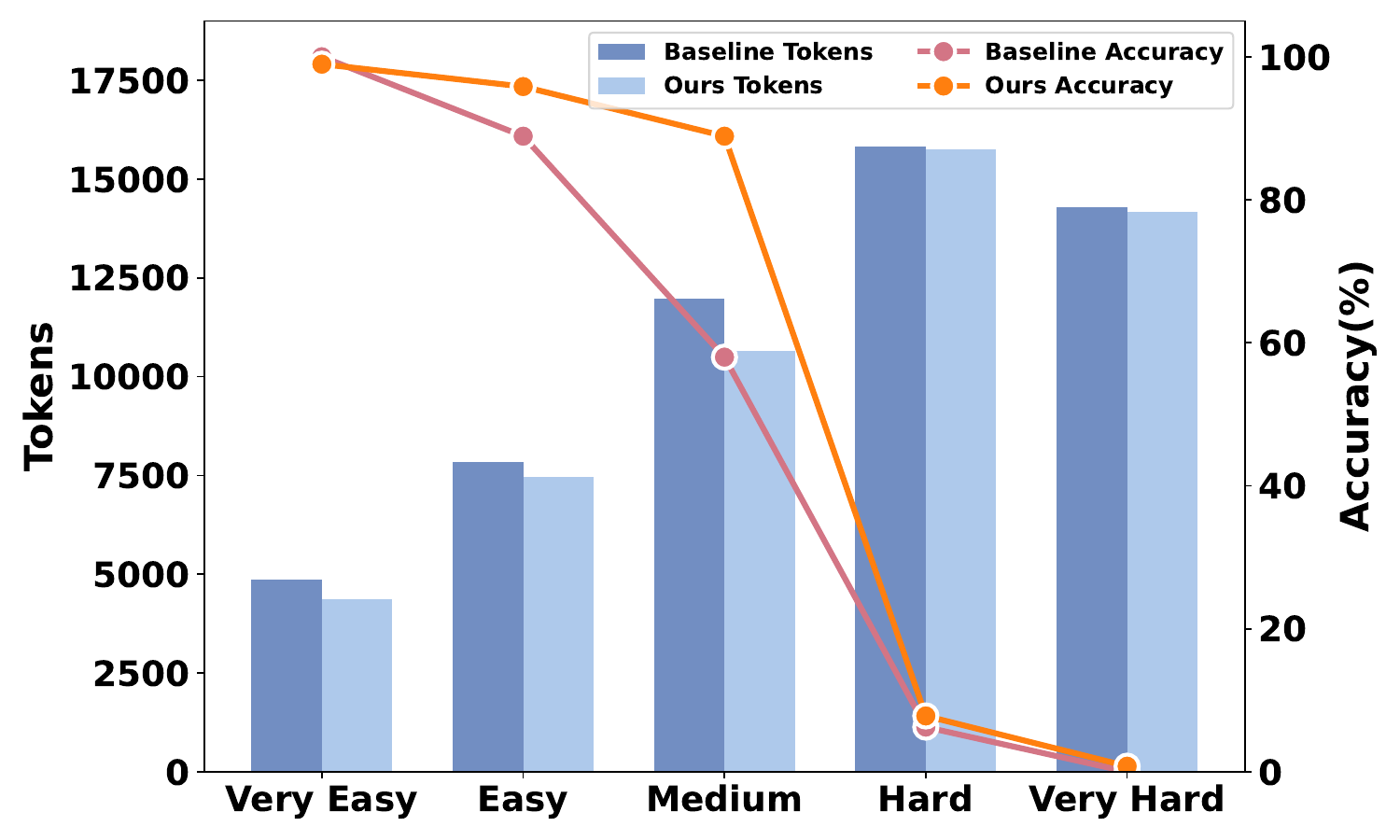}
        \caption{DeepSeek-R1-Distill-Qwen-14B}
        \label{fig:difficulty_leap_i_14B}
    \end{subfigure}
    \caption{The Pass@1 and average token distribution across different difficulty levels, from Very Easy to Very Hard, for the 7B and 14B versions of DeepSeek-R1-Distill-Qwen.}
    \label{fig:difficulty_leap_i_main} \vspace{-3mm}
\end{figure}

To answer this question, we conduct a detailed analysis across different difficulty levels.
Specifically, for each model, we first divide questions in AIME 2024 into five difficulty levels by counting correct responses in the baseline (independent 32 parallel reasoning without peers).
The difficulty levels are as follows: ``Very Easy'' corresponds to 32 correct responses, ``Easy'' includes 25 to 31 correct, ``Medium'' consists of 9 to 24 correct responses, ``Hard'' includes 1 to 8 correct, and ``Very Hard'' refers to 0 correct responses.
Then, we compute accuracy and token usage under each level.
To better analyze tokens used solely for reasoning, we exclude summaries in the LeaP blocks collected from other paths.
As shown in Figure~\ref{fig:difficulty_leap_i_main}, we report results for DeepSeek-Distill-Qwen-7B and DeepSeek-Distill-Qwen-14B.
We find a consistent trend: \textbf{LeaP improves accuracy across all difficulty levels}, including those where the baseline model completely fails.
This result is surprising and encouraging.
Even for ``Very Hard'' questions—where the baseline fails to solve a single instance—LeaP achieves meaningful gains.
This suggests that LeaP not only amplifies partial correctness but can also help recover from complete failure, enabling models to solve problems that were previously unreachable.
In addition, after removing tokens attributed to peer summaries, LeaP often consumes fewer tokens than the baseline.
Combined with our earlier finding in Section~\ref{sec:eval_results} that LeaP reduces the number of ``Aha'' moments, this supports the view that LeaP helps the reasoning model reach consensus earlier, thereby reducing overthinking.
We include additional results for the LeaP-T series in Appendix~\ref{app:leap_t_diff}, which further confirm the robustness of LeaP across varying levels of difficulty.

\subsection{Human Verification on LeaP}\label{sec:analysis_human}
\begin{wrapfigure}{R}{0.4\textwidth}
    \centering \vspace{-6mm}
    \includegraphics[width=\linewidth]{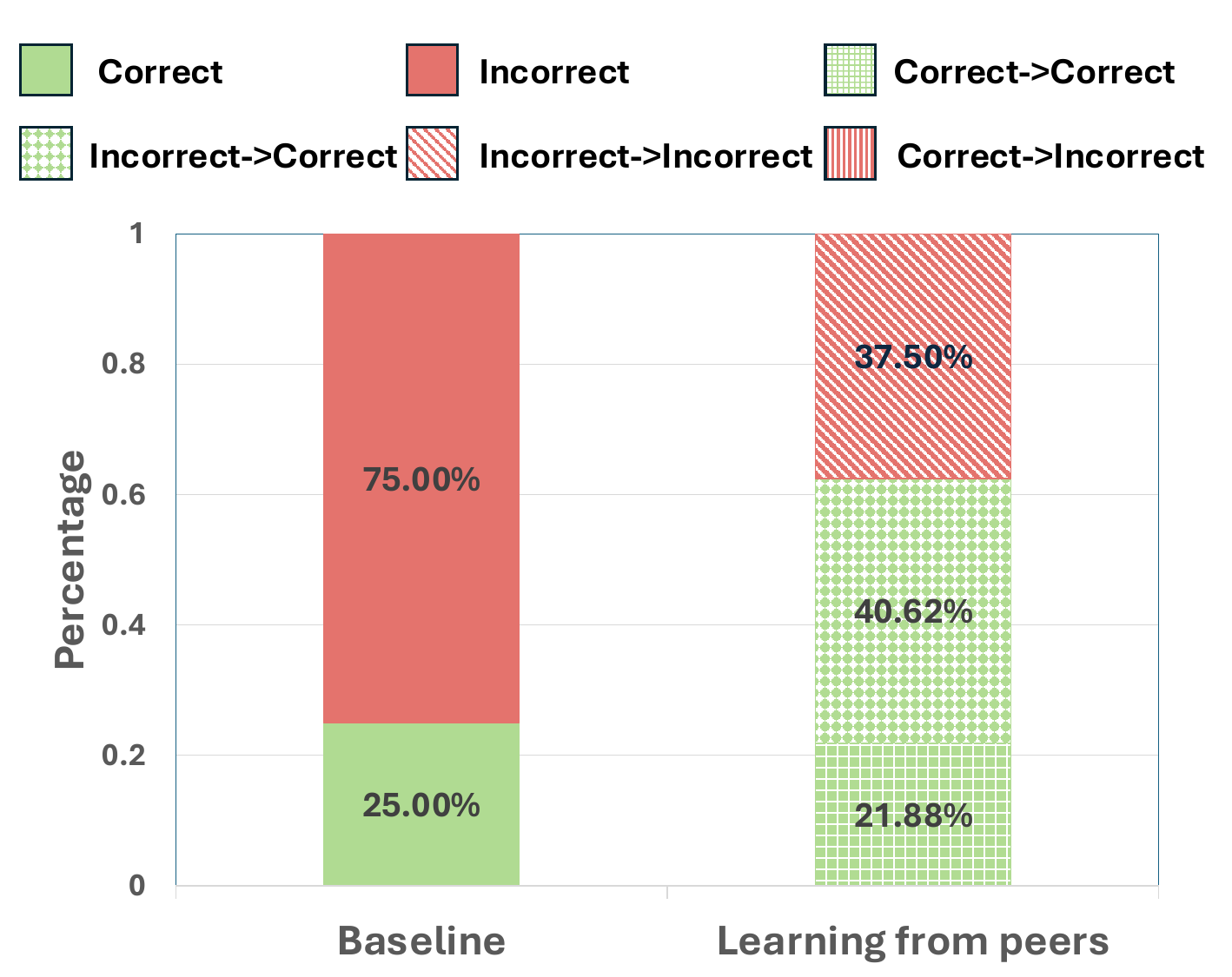}
    \caption{Case study comparing QwQ-32B responses with and without LeaP on the 11\textsuperscript{th} problem of AIME 2024.}
    \label{fig:case_study} \vspace{-6mm}
\end{wrapfigure}
While results across various benchmarks provide consistent evidence that LeaP significantly improves reasoning ability, they do not always reveal underlying behavioral changes induced by LeaP.
To gain deeper insight, we perform a detailed human evaluation by comparing QwQ-32B outputs with and without LeaP on a specific example—the 11\textsuperscript{th} problem from AIME 2024.
As shown in Figure~\ref{fig:case_study}, the baseline model (without LeaP) produces correct answers in only 25.00\% of cases (8 out of 32).
In contrast, when using LeaP, the proportion of correct responses rises significantly to 62.50\% (20 out of 32), confirming the substantial benefit of learning from peers.
To better understand the source of this improvement, we categorize responses into four types: \textit{Correct~$\rightarrow$~Correct}, \textit{Incorrect~$\rightarrow$~Correct}, \textit{Incorrect~$\rightarrow$~Incorrect}, and \textit{Correct~$\rightarrow$~Incorrect}.
Remarkably, we observe that 40.62\% of responses fall into the \textit{Incorrect~$\rightarrow$~Correct} category.
This indicates that many reasoning paths are initially flawed but are successfully repaired after communicating with other peers.
Even more telling, no responses fall into the \textit{Correct~$\rightarrow$~Incorrect} case, suggesting that peer communication rarely disrupts already correct reasoning.
Instead, it primarily serves as a corrective mechanism that identifies and fixes errors. Case studies are provided in Appendix~\ref{app:case_study_success}

\section{Related Work}

\subsection{Self-Correction Bottleneck}
Recent studies show that while current Large Reasoning Models can generate lengthy reasoning chains, their performance does not consistently improve with increasing chain length~\cite{zeng2025revisiting}. In fact, longer reasoning chains often result in lower accuracy, which contradicts the assumption that extended reasoning improves problem-solving capabilities.
Recent study~\cite{marjanovic2025deepseek} further investigates the reasoning behavior of DeepSeek-R1, highlighting its tendency to dwell on previously explored problem formulations. This behavior hinders further exploration and leads to suboptimal reasoning paths.
In this paper, we reaffirm their findings and present an effective method to address this issue.

\subsection{(Interactive) Parallel Inference in LLMs}
Parallel inference~\cite{zeng2024scaling, zhang2025and, snell2024scaling} enables LLMs to generate multiple reasoning paths simultaneously and aggregate them into a final answer.
Self-Consistency~(Majority Voting)~\cite{zeng2025revisiting, snell2024scaling, wang2022self, li2023making, brown2024large, song2024good, nguyen2024consistent} selects the final answer by voting over candidate responses and choosing the one with the highest number of votes.
Best-of-$N$~\cite{gao2023scaling, cobbe2021training, sun2024fast, gui2024bonbon, amini2024variational, sessa2024bond} improves response quality by generating multiple candidates and selecting the one with the highest reward.
Another important direction involves interaction among multiple LLMs within collaborative frameworks during parallel inference.
Multi-agent debate~\cite{du2023improving} establishes a symmetric discussion mechanism among agents.
ReConcile~\cite{chen2023reconcile} and Corex~\cite{sun2023corex} treat collaboration as multi-round discussions or deliberations, using consensus mechanisms and role specialization (e.g., proposer, reviewer) to improve answer reliability.
Building on these ideas, methods such as CoMM~\cite{chen2024comm} and MALT~\cite{motwani2024malt} introduce explicit agent roles and diverse reasoning paths, enabling joint training for complex tasks.
MoA~\cite{wang2024mixture, li2024smoa} further proposes architectural hierarchies and network-based communication patterns to enhance collective reasoning.
The key difference between our approach and methods like MoA lies in the \textbf{interaction mechanism between reasoning paths}.
In MoA, each round only accesses the output from the previous round, without reference to earlier context.
In contrast, our method enables \textbf{direct collaboration across multiple reasoning paths}, maintaining a finer-grained and more complete history of the reasoning process.

\section{Conclusion and Future Work}
In this work, we identify a phenomenon we term the \emph{``Prefix Dominance Trap''}, which reveals that the self-correction ability in large reasoning models (LRMs) is still limited.
To address this issue, we introduce a novel approach to enhance their performance through Learning from Peers (LeaP).
Intuitively, rather than a closed-book exam, this method facilitates note-passing among reasoning path peers, allowing them to borrow insights from one another.
We further develop a series of trained models to empower LeaP, referred to as LeaP-T.
Our experiments show that LeaP significantly improves models' ability to recover from initial errors, often outperforming baselines and even larger models in certain cases.
Our in-depth analysis further uncovers the mechanisms behind these improvements, offering valuable insights into overcoming the self-correction limitations in LRMs.
Overall, our findings highlight the potential of leveraging peer collaboration to improve the reasoning abilities of large models.

We also propose two promising directions for future work:
\begin{itemize}
    \item \textbf{Learning from peers in Reinforcement Learning:} 
    Extending peer learning to reinforcement learning (RL) is an exciting direction. By incorporating LeaP into RL, models could learn to collaborate more effectively in solving complex problems, potentially unlocking the full capabilities of base models.
    \item \textbf{Learning from peers with different expertise:} 
    Another avenue is to leverage peers with specialized expertise, using different prompts and tools. For example, one group of peers could use web searches while another employs Python for problem-solving. This could enhance reasoning quality, particularly in multi-faceted tasks.
\end{itemize}

\bibliographystyle{unsrt}
\bibliography{neurips_2025}

\begin{thebibliography}{10}

\bibitem{openai2024a}
{OpenAI}.
\newblock Learning to reason with llms, 2024.
\newblock Accessed: 2025-05-03.

\bibitem{openai2024b}
{OpenAI}.
\newblock Openai o1 system card, 2024.
\newblock Accessed: 2025-05-03.

\bibitem{openai2025o3}
{OpenAI}.
\newblock Openai o3 mini, 2025.
\newblock Accessed: 2025-05-03.

\bibitem{guo2025deepseek}
Daya Guo, Dejian Yang, Haowei Zhang, Junxiao Song, Ruoyu Zhang, Runxin Xu, Qihao Zhu, Shirong Ma, Peiyi Wang, Xiao Bi, et~al.
\newblock Deepseek-r1: Incentivizing reasoning capability in llms via reinforcement learning.
\newblock {\em arXiv preprint arXiv:2501.12948}, 2025.

\bibitem{qwq32b}
Qwen Team.
\newblock Qwq-32b: Embracing the power of reinforcement learning, March 2025.

\bibitem{zhang2025and}
Qiyuan Zhang, Fuyuan Lyu, Zexu Sun, Lei Wang, Weixu Zhang, Zhihan Guo, Yufei Wang, Irwin King, Xue Liu, and Chen Ma.
\newblock What, how, where, and how well? a survey on test-time scaling in large language models.
\newblock {\em arXiv preprint arXiv:2503.24235}, 2025.

\bibitem{snell2024scaling}
Charlie Snell, Jaehoon Lee, Kelvin Xu, and Aviral Kumar.
\newblock Scaling llm test-time compute optimally can be more effective than scaling model parameters.
\newblock {\em arXiv preprint arXiv:2408.03314}, 2024.

\bibitem{zeng2025revisiting}
Zhiyuan Zeng, Qinyuan Cheng, Zhangyue Yin, Yunhua Zhou, and Xipeng Qiu.
\newblock Revisiting the test-time scaling of o1-like models: Do they truly possess test-time scaling capabilities?
\newblock {\em arXiv preprint arXiv:2502.12215}, 2025.

\bibitem{chen2024expanding}
Zhe Chen, Weiyun Wang, Yue Cao, Yangzhou Liu, Zhangwei Gao, Erfei Cui, Jinguo Zhu, Shenglong Ye, Hao Tian, Zhaoyang Liu, et~al.
\newblock Expanding performance boundaries of open-source multimodal models with model, data, and test-time scaling.
\newblock {\em arXiv preprint arXiv:2412.05271}, 2024.

\bibitem{giuliodori2006peer}
Mauricio~J Giuliodori, Heidi~L Lujan, and Stephen~E DiCarlo.
\newblock Peer instruction enhanced student performance on qualitative problem-solving questions.
\newblock {\em Advances in physiology education}, 30(4):168--173, 2006.

\bibitem{snyder2015implementation}
Julia~J Snyder, B~Elijah Carter, and Jason~R Wiles.
\newblock Implementation of the peer-led team-learning instructional model as a stopgap measure improves student achievement for students opting out of laboratory.
\newblock {\em CBE—Life Sciences Education}, 14(1):ar2, 2015.

\bibitem{falk2006clean}
Armin Falk and Andrea Ichino.
\newblock Clean evidence on peer effects.
\newblock {\em Journal of labor economics}, 24(1):39--57, 2006.

\bibitem{aime2024}
{MAA}.
\newblock {American Invitational Mathematics Examination - AIME 2024}, February 2024.
\newblock \url{https://maa.org/math-competitions/american-invitational-mathematics-examination-aime}.

\bibitem{aime2025}
{MAA}.
\newblock {American Invitational Mathematics Examination - AIME 2025}, February 2025.
\newblock \url{https://maa.org/math-competitions/american-invitational-mathematics-examination-aime}.

\bibitem{ai-mathematical-olympiad-progress-prize-2}
Simon Frieder, Sam Bealing, Arsenii Nikolaiev, Geoff~C. Smith, Kevin Buzzard, Timothy Gowers, Peter~J. Liu, Po-Shen Loh, Lester Mackey, Leonardo de~Moura, Dan Roberts, D.~Sculley, Terence Tao, David Balduzzi, Simon Coyle, Alex Gerko, Ryan Holbrook, Addison Howard, and XTX Markets.
\newblock Ai mathematical olympiad - progress prize 2.
\newblock \url{https://kaggle.com/competitions/ai-mathematical-olympiad-progress-prize-2}, 2024.
\newblock Kaggle.

\bibitem{rein2024gpqa}
David Rein, Betty~Li Hou, Asa~Cooper Stickland, Jackson Petty, Richard~Yuanzhe Pang, Julien Dirani, Julian Michael, and Samuel~R. Bowman.
\newblock {GPQA}: A graduate-level google-proof q\&a benchmark.
\newblock In {\em First Conference on Language Modeling}, 2024.

\bibitem{levenshtein1966binary}
Vladimir~I Levenshtein et~al.
\newblock Binary codes capable of correcting deletions, insertions, and reversals.
\newblock In {\em Soviet physics doklady}, volume~10, pages 707--710. Soviet Union, 1966.

\bibitem{kwon2023efficient}
Woosuk Kwon, Zhuohan Li, Siyuan Zhuang, Ying Sheng, Lianmin Zheng, Cody~Hao Yu, Joseph~E. Gonzalez, Hao Zhang, and Ion Stoica.
\newblock Efficient memory management for large language model serving with pagedattention.
\newblock In {\em Proceedings of the ACM SIGOPS 29th Symposium on Operating Systems Principles}, 2023.

\bibitem{wang2022self}
Xuezhi Wang, Jason Wei, Dale Schuurmans, Quoc Le, Ed~Chi, Sharan Narang, Aakanksha Chowdhery, and Denny Zhou.
\newblock Self-consistency improves chain of thought reasoning in language models.
\newblock {\em arXiv preprint arXiv:2203.11171}, 2022.

\bibitem{zeng2025simplerl}
Weihao Zeng, Yuzhen Huang, Qian Liu, Wei Liu, Keqing He, Zejun Ma, and Junxian He.
\newblock Simplerl-zoo: Investigating and taming zero reinforcement learning for open base models in the wild.
\newblock {\em arXiv preprint arXiv:2503.18892}, 2025.

\bibitem{li2025start}
Chengpeng Li, Mingfeng Xue, Zhenru Zhang, Jiaxi Yang, Beichen Zhang, Xiang Wang, Bowen Yu, Binyuan Hui, Junyang Lin, and Dayiheng Liu.
\newblock Start: Self-taught reasoner with tools.
\newblock {\em arXiv preprint arXiv:2503.04625}, 2025.

\bibitem{Slow_Thinking_with_LLMs_3}
Zhipeng Chen, Yingqian Min, Beichen Zhang, Jie Chen, Jinhao Jiang, Daixuan Cheng, Wayne~Xin Zhao, Zheng Liu, Xu~Miao, Yang Lu, Lei Fang, Zhongyuan Wang, and Ji-Rong Wen.
\newblock An empirical study on eliciting and improving r1-like reasoning models.
\newblock {\em arXiv preprint arXiv:2503.04548}, 2025.

\bibitem{wang2024mixture}
Junlin Wang, Jue Wang, Ben Athiwaratkun, Ce~Zhang, and James Zou.
\newblock Mixture-of-agents enhances large language model capabilities.
\newblock {\em arXiv preprint arXiv:2406.04692}, 2024.

\bibitem{openai2024gpt4o}
{OpenAI}.
\newblock Hello gpt-4o.
\newblock \url{https://openai.com/index/hello-gpt-4o/}, 2024.
\newblock Accessed: 2025-05-03.

\bibitem{marjanovic2025deepseek}
Sara~Vera Marjanovi{\'c}, Arkil Patel, Vaibhav Adlakha, Milad Aghajohari, Parishad BehnamGhader, Mehar Bhatia, Aditi Khandelwal, Austin Kraft, Benno Krojer, Xing~Han L{\`u}, et~al.
\newblock Deepseek-r1 thoughtology: Let's< think> about llm reasoning.
\newblock {\em arXiv preprint arXiv:2504.07128}, 2025.

\bibitem{zeng2024scaling}
Zhiyuan Zeng, Qinyuan Cheng, Zhangyue Yin, Bo~Wang, Shimin Li, Yunhua Zhou, Qipeng Guo, Xuanjing Huang, and Xipeng Qiu.
\newblock Scaling of search and learning: A roadmap to reproduce o1 from reinforcement learning perspective.
\newblock {\em arXiv preprint arXiv:2412.14135}, 2024.

\bibitem{li2023making}
Yifei Li, Zeqi Lin, Shizhuo Zhang, Qiang Fu, Bei Chen, Jian-Guang Lou, and Weizhu Chen.
\newblock Making language models better reasoners with step-aware verifier.
\newblock In {\em Proceedings of the 61st Annual Meeting of the Association for Computational Linguistics (Volume 1: Long Papers)}, pages 5315--5333, 2023.

\bibitem{brown2024large}
Bradley Brown, Jordan Juravsky, Ryan Ehrlich, Ronald Clark, Quoc~V Le, Christopher R{\'e}, and Azalia Mirhoseini.
\newblock Large language monkeys: Scaling inference compute with repeated sampling.
\newblock {\em arXiv preprint arXiv:2407.21787}, 2024.

\bibitem{song2024good}
Yifan Song, Guoyin Wang, Sujian Li, and Bill~Yuchen Lin.
\newblock The good, the bad, and the greedy: Evaluation of llms should not ignore non-determinism.
\newblock {\em arXiv preprint arXiv:2407.10457}, 2024.

\bibitem{nguyen2024consistent}
Alex Nguyen, Dheeraj Mekala, Chengyu Dong, and Jingbo Shang.
\newblock When is the consistent prediction likely to be a correct prediction?
\newblock {\em arXiv preprint arXiv:2407.05778}, 2024.

\bibitem{gao2023scaling}
Leo Gao, John Schulman, and Jacob Hilton.
\newblock Scaling laws for reward model overoptimization.
\newblock In {\em International Conference on Machine Learning}, pages 10835--10866. PMLR, 2023.

\bibitem{cobbe2021training}
Karl Cobbe, Vineet Kosaraju, Mohammad Bavarian, Mark Chen, Heewoo Jun, Lukasz Kaiser, Matthias Plappert, Jerry Tworek, Jacob Hilton, Reiichiro Nakano, et~al.
\newblock Training verifiers to solve math word problems.
\newblock {\em arXiv preprint arXiv:2110.14168}, 2021.

\bibitem{sun2024fast}
Hanshi Sun, Momin Haider, Ruiqi Zhang, Huitao Yang, Jiahao Qiu, Ming Yin, Mengdi Wang, Peter Bartlett, and Andrea Zanette.
\newblock Fast best-of-n decoding via speculative rejection.
\newblock {\em arXiv preprint arXiv:2410.20290}, 2024.

\bibitem{gui2024bonbon}
Lin Gui, Cristina G{\^a}rbacea, and Victor Veitch.
\newblock Bonbon alignment for large language models and the sweetness of best-of-n sampling.
\newblock {\em arXiv preprint arXiv:2406.00832}, 2024.

\bibitem{amini2024variational}
Afra Amini, Tim Vieira, Elliott Ash, and Ryan Cotterell.
\newblock Variational best-of-n alignment.
\newblock {\em arXiv preprint arXiv:2407.06057}, 2024.

\bibitem{sessa2024bond}
Pier~Giuseppe Sessa, Robert Dadashi, L{\'e}onard Hussenot, Johan Ferret, Nino Vieillard, Alexandre Ram{\'e}, Bobak Shariari, Sarah Perrin, Abe Friesen, Geoffrey Cideron, et~al.
\newblock Bond: Aligning llms with best-of-n distillation.
\newblock {\em arXiv preprint arXiv:2407.14622}, 2024.

\bibitem{du2023improving}
Yilun Du, Shuang Li, Antonio Torralba, Joshua~B Tenenbaum, and Igor Mordatch.
\newblock Improving factuality and reasoning in language models through multiagent debate.
\newblock In {\em Forty-first International Conference on Machine Learning}, 2023.

\bibitem{chen2023reconcile}
Justin Chih-Yao Chen, Swarnadeep Saha, and Mohit Bansal.
\newblock Reconcile: Round-table conference improves reasoning via consensus among diverse llms.
\newblock {\em arXiv preprint arXiv:2309.13007}, 2023.

\bibitem{sun2023corex}
Qiushi Sun, Zhangyue Yin, Xiang Li, Zhiyong Wu, Xipeng Qiu, and Lingpeng Kong.
\newblock Corex: Pushing the boundaries of complex reasoning through multi-model collaboration.
\newblock {\em arXiv preprint arXiv:2310.00280}, 2023.

\bibitem{chen2024comm}
Pei Chen, Boran Han, and Shuai Zhang.
\newblock Comm: Collaborative multi-agent, multi-reasoning-path prompting for complex problem solving.
\newblock {\em arXiv preprint arXiv:2404.17729}, 2024.

\bibitem{motwani2024malt}
Sumeet~Ramesh Motwani, Chandler Smith, Rocktim~Jyoti Das, Rafael Rafailov, Ivan Laptev, Philip~HS Torr, Fabio Pizzati, Ronald Clark, and Christian~Schroeder de~Witt.
\newblock Malt: Improving reasoning with multi-agent llm training.
\newblock {\em arXiv preprint arXiv:2412.01928}, 2024.

\bibitem{li2024smoa}
Dawei Li, Zhen Tan, Peijia Qian, Yifan Li, Kumar~Satvik Chaudhary, Lijie Hu, and Jiayi Shen.
\newblock Smoa: Improving multi-agent large language models with sparse mixture-of-agents.
\newblock {\em arXiv preprint arXiv:2411.03284}, 2024.

\bibitem{ji2025first}
Ke~Ji, Jiahao Xu, Tian Liang, Qiuzhi Liu, Zhiwei He, Xingyu Chen, Xiaoyuan Liu, Zhijie Wang, Junying Chen, Benyou Wang, et~al.
\newblock The first few tokens are all you need: An efficient and effective unsupervised prefix fine-tuning method for reasoning models.
\newblock {\em arXiv preprint arXiv:2503.02875}, 2025.

\end{thebibliography}

\newpage
\appendix

\section{Prompts}\label{app:prompts}

\begin{figure}[H]
\centering
\begin{prompt}{Summarization Prompt Structure}
\textbf{\textcolor{cyan}{Summary Trigger:}} (randomly select one)

\begin{itemize}
    \item Alright, let's take a step back and summarize what we've figured out so far.
    \item Wait, let me quickly recap what I've concluded so far.
    \item Alright, let me shortly review the conclusions I've drawn so I can move forward more efficiently.
    \item Hmm, a quick summary of what I've figured out might help streamline the next part of my reasoning.
    \item Hold on, I should summarize the key points briefly to ensure I'm on the right track.
    \item Okay, before continuing, let me put together a brief summary of the insights I've gathered so far.
    \item Okay, time to consolidate everything I've found into a concise summary.
\end{itemize}

\vspace{0.5em}
\textbf{\textcolor{orange}{Summary Template:}} (randomly select one)

\begin{itemize}
    \item In short, my current conclusions are that ...
    \item To summarize, based on my previous reasoning, I have currently found that ...
    \item In conclusion, the current key takeaways and results are ...
    \item In short, I've currently concluded that ...
    \item To summarize, my recent findings are ...
    \item In conclusion, the current insights and results I've gathered are ...
\end{itemize}
\end{prompt}
\caption{The structure of the summarization prompt used during LeaP.}
\label{fig:prompt_summarize}
\end{figure}

\begin{figure}[H]
\centering
\begin{prompt}{Prompt for Different Tasks}
\textcolor{blue}{GPQA:}

Please show your choice in the answer field with only the choice letter, e.g.,``ANSWER'': ``C''.

\textcolor{blue}{Math Tasks:}

Please reason step by step, and put your final answer within $\backslash$boxed{}.

\end{prompt}
\caption{Prompts for different tasks.}
\label{fig:task_prompts}
\end{figure}

\begin{figure}[H]
\centering
\begin{prompt}{Prompt for Mixture-of-Agents}
Problem: \{problem\}

You have been provided with a set of responses from various open-source models to the latest user query. Your task is to synthesize these responses into a single, high-quality response. It is crucial to critically evaluate the information provided in these responses, recognizing that some of it may be biased or incorrect. Your response should not simply replicate the given answers but should offer a refined, accurate, and comprehensive reply to the instruction. Ensure your response is well-structured, coherent, and adheres to the highest standards of accuracy and reliability.

Responses from models:
\end{prompt}
\caption{Prompts for mixture of agents.}
\label{fig:prompt_moa}
\end{figure}

\section{Prefix Dominance Trap for Starting with Good Beginnings}\label{app:good_beginning}

While the ``Prefix Dominance Trap'' highlights how poor beginnings can severely constrain reasoning, a natural follow-up question is whether good beginnings are sufficient to ensure correct final answers.
To this end, we conduct a follow-up experiment similar to those in Section~\ref{sec:motivation} under the \textit{good beginnings} setting, where the model starts reasoning with a beginning of 15\% average response length from correct reasoning paths.

\begin{figure}[H]
\centering
  \begin{subfigure}[b]{0.5\textwidth}
        \centering
        \includegraphics[width=\linewidth]{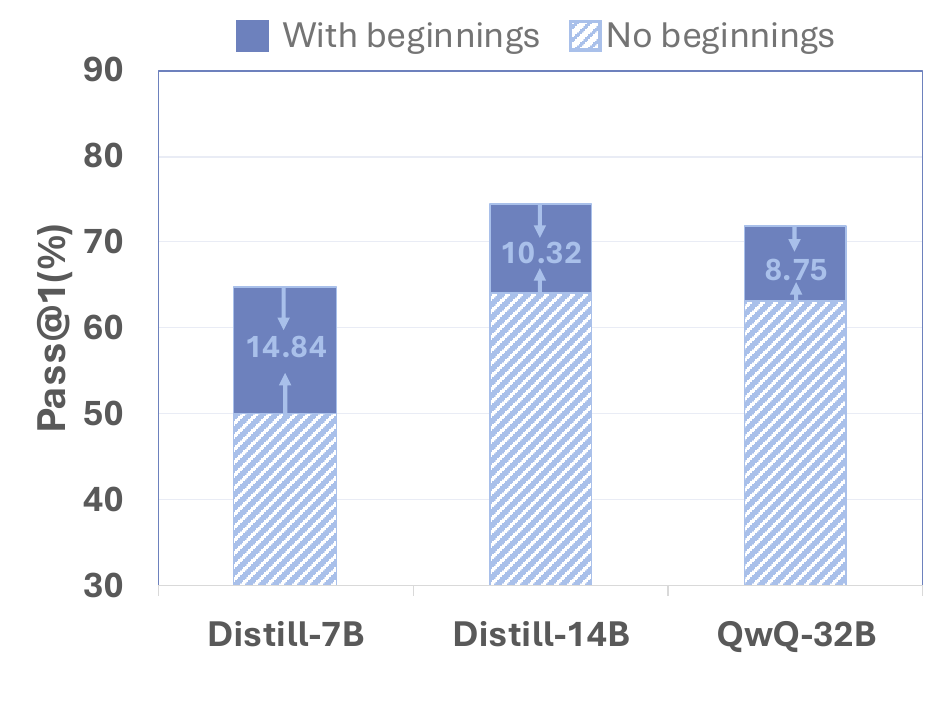}
        \caption{Independent Reasoning}
        \label{fig:motivation_baseline_correct}
    \end{subfigure}
  \hspace{-2mm}
  \begin{subfigure}[b]{0.5\textwidth}
        \centering
        \includegraphics[width=\linewidth]{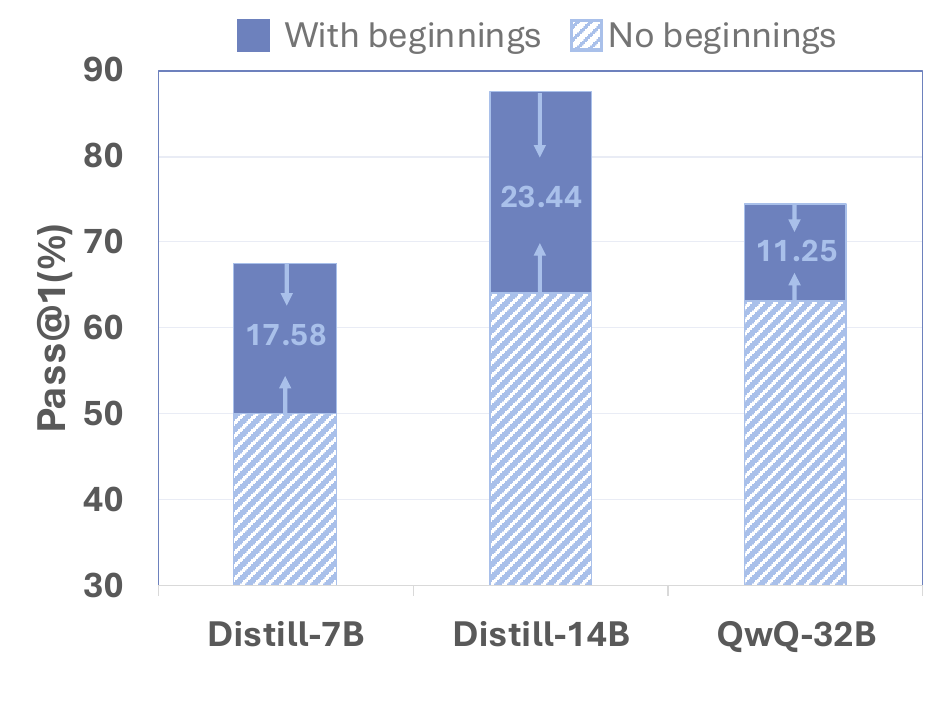}
        \caption{LeaP}
        \label{fig:motivation_ours_correct}
    \end{subfigure}
    \caption{The results of starting with a good beginning of 15\% average response length.}
    \label{fig:good_beginning}
\end{figure}

As shown in Figure~\ref{fig:motivation_baseline_correct}, models initialized with good beginnings indeed perform better than random or bad initializations.
This suggests that early correct cues provide useful guidance.
However, these beginnings do not fully eliminate reasoning errors—many responses still deviate in later steps and arrive at incorrect conclusions.
This observation reinforces that good beginnings can reduce—but not entirely prevent—reasoning failures.

We then examine how LeaP performs in the same setting.
As illustrated in Figure~\ref{fig:motivation_ours_correct}, our method consistently surpasses independent reasoning, even when starting from already high-quality beginnings.
For instance, DeepSeek-Distill-Qwen-14B improves from 74.38 to 87.50 in Pass@1.
This significant gain indicates that LeaP not only repairs faulty reasoning from poor prefixes but also mitigates subtle errors that emerge even when the reasoning starts correctly.

\section{Additional Evaluation Results}\label{app:eval}

\subsection{Random Routing Results}

\begin{table}[H]
    \centering
    \small
    \caption{The results of Pass@1 for four routing mechanisms on four benchmarks.}
    \begin{tabular}{lccccccccc}
    \toprule
    \multirow{2}{*}{\textbf{Benchmarks}} &\multirow{2}{*}{\textbf{Models}} & \multicolumn{2}{c}{\textbf{Random}} &\multicolumn{2}{c}{\textbf{Clustered}}& \multicolumn{2}{c}{\textbf{Hybrid}} & \multicolumn{2}{c}{\textbf{Dispersed}} \\
    
    \cline{3-10}
    & & \textbf{Top-2} & \textbf{Top-4} & \textbf{Top-2} & \textbf{Top-4}& \textbf{Top-2} & \textbf{Top-4} &  \textbf{Top-2} & \textbf{Top-4}   \\
    
    \specialrule{0.5pt}{2pt}{2pt}
    \multirow{3}{*}{AIME 2024} 
    & R1-7B   & 
    \cellcolor{mygreen2}58.65 & 
    \cellcolor{mygreen3}\textcolor{white}{60.83} & 
    \cellcolor{mygreen1}56.15 &
    \cellcolor{mygreen3}\textcolor{white}{59.27} &
    \cellcolor{mygreen2}59.17 &
    \cellcolor{mygreen4}\textcolor{white}{61.67}&
    \cellcolor{mygreen3}\textcolor{white}{60.31} &
    \cellcolor{mygreen3}\textcolor{white}{60.52} 
    \\
    & R1-14B  & 
     \cellcolor{mygreen2}72.60 & 
    \cellcolor{mygreen3}\textcolor{white}{75.83} & 
    \cellcolor{mygreen2}72.08&
    \cellcolor{mygreen1}71.04  &

    \cellcolor{mygreen2}71.77 &
    \cellcolor{mygreen3}\textcolor{white}{74.48} &
    \cellcolor{mygreen2}71.15 & 
    \cellcolor{mygreen4}\textcolor{white}{77.29}  
    \\
    & QwQ-32B &
    \cellcolor{mygreen2}80.21 & 
    \cellcolor{mygreen3}\textcolor{white}{81.67} & 
    \cellcolor{mygreen1}78.96&
    \cellcolor{mygreen3}\textcolor{white}{81.56} &

    \cellcolor{mygreen3}\textcolor{white}{81.88} &
    \cellcolor{mygreen3}\textcolor{white}{81.67}&
    \cellcolor{mygreen3}\textcolor{white}{81.56} &
    \cellcolor{mygreen4}\textcolor{white}{85.83} \\

    \specialrule{0.5pt}{2pt}{2pt}
    \multirow{3}{*}{AIMO 2025} 
    & R1-7B   & 
    \cellcolor{mygreen2}40.62 &
    \cellcolor{mygreen3}\textcolor{white}{42.81}  & 
    \cellcolor{mygreen1}37.81  &
    \cellcolor{mygreen2}39.06  &

    \cellcolor{mygreen3}\textcolor{white}{41.88}  &
    \cellcolor{mygreen2}40.31  &
    \cellcolor{mygreen2}39.06  & 
    \cellcolor{mygreen4}\textcolor{white}{45.00}  

    \\
    & R1-14B  &
    \cellcolor{mygreen3}\textcolor{white}{52.50} &
    \cellcolor{mygreen3}\textcolor{white}{53.13} &
    \cellcolor{mygreen2}51.50&
    \cellcolor{mygreen4}\textcolor{white}{55.31}&

    \cellcolor{mygreen2}{51.88} & 
    \cellcolor{mygreen3}\textcolor{white}{52.19}&
    \cellcolor{mygreen1}49.38 &
    \cellcolor{mygreen2}{51.25} 
    \\
    & QwQ-32B &
    \cellcolor{mygreen2}63.75 &
    \cellcolor{mygreen4}\textcolor{white}{68.44} &
    \cellcolor{mygreen1}63.75&
    \cellcolor{mygreen3}\textcolor{white}{65.63} &

    \cellcolor{mygreen2}64.06  &
    \cellcolor{mygreen2}64.69&
    \cellcolor{mygreen4}\textcolor{white}{67.19} &
    \cellcolor{mygreen4}\textcolor{white}{67.19} 
    \\
    
    \specialrule{0.5pt}{2pt}{2pt}
    \multirow{3}{*}{AIME 2025} 
    & R1-7B    & 
    \cellcolor{mygreen2}38.02   &
    \cellcolor{mygreen3}\textcolor{white}{39.17}   &
    \cellcolor{mygreen1}36.98&
    \cellcolor{mygreen2}{37.50}&

    \cellcolor{mygreen2}37.19  &
    \cellcolor{mygreen3}\textcolor{white}{39.27}&
        \cellcolor{mygreen4}\textcolor{white}{40.93}  &
    \cellcolor{mygreen3}\textcolor{white}{38.44}  

    \\
    & R1-14B   &
    \cellcolor{mygreen1}48.96   &
    \cellcolor{mygreen3}\textcolor{white}{52.29}   &
    \cellcolor{mygreen2}50.42&
    \cellcolor{mygreen2}{51.88}&

    \cellcolor{mygreen4}\textcolor{white}{54.38}  &
    \cellcolor{mygreen4}\textcolor{white}{54.38}&
        \cellcolor{mygreen2}50.31 &
    \cellcolor{mygreen4}\textcolor{white}{54.17}  
    \\
    & QwQ-32B  &
     \cellcolor{mygreen3}\textcolor{white}{71.35}   &
    \cellcolor{mygreen3}\textcolor{white}{71.98}   &
    \cellcolor{mygreen1}{68.85} &
    \cellcolor{mygreen3}\textcolor{white}{71.35} &

    \cellcolor{mygreen3}\textcolor{white}{71.04}  &
    \cellcolor{mygreen4}\textcolor{white}{72.50}&
        \cellcolor{mygreen2}{70.83} &
    \cellcolor{mygreen3}\textcolor{white}{71.67}  
    \\

    \specialrule{0.5pt}{2pt}{2pt}
    \multirow{2}{*}{GPQA} 
    & R1-7B   &
    \cellcolor{mygreen2}52.96 &
    \cellcolor{mygreen2}52.65 &
    \cellcolor{mygreen2}52.97&
    \cellcolor{mygreen2}52.65&

    \cellcolor{mygreen1}51.83 &
    \cellcolor{mygreen3}\textcolor{white}{53.47}&
        \cellcolor{mygreen3}\textcolor{white}{53.28} &
    \cellcolor{mygreen4}\textcolor{white}{55.56} \\

    \multirow{2}{*}{Diamond} & R1-14B  &
    \cellcolor{mygreen2}54.99 &
    \cellcolor{mygreen3}\textcolor{white}{57.26} &
    \cellcolor{mygreen2}54.80&
    \cellcolor{mygreen4}\textcolor{white}{58.33}&

    \cellcolor{mygreen1}54.42 &
    \cellcolor{mygreen3}\textcolor{white}{57.89}&
        \cellcolor{mygreen3}\textcolor{white}{55.68} &
    \cellcolor{mygreen2}55.05 \\
    
    & QwQ-32B &
    \cellcolor{mygreen3}\textcolor{white}{64.65} &
    \cellcolor{mygreen3}\textcolor{white}{64.89} &
    \cellcolor{mygreen3}\textcolor{white}{65.03}&
    \cellcolor{mygreen3}\textcolor{white}{65.28}&

    \cellcolor{mygreen1}61.87 &
    \cellcolor{mygreen4}\textcolor{white}{65.21}&
        \cellcolor{mygreen4}\textcolor{white}{66.16} &
    \cellcolor{mygreen2}63.32\\
    
    \specialrule{0.5pt}{2pt}{2pt}
    \multirow{3}{*}{Avg.} 
    & R1-7B  &
    \cellcolor{mygreen2}47.56&
     \cellcolor{mygreen3}\textcolor{white}{48.87}&
    \cellcolor{mygreen1}45.98& 
    \cellcolor{mygreen2}{47.12}&

    \cellcolor{mygreen2}{47.52}&	
    \cellcolor{mygreen3}\textcolor{white}{48.68}&
        \cellcolor{mygreen3}\textcolor{white}{48.40}&	
    \cellcolor{mygreen4}\textcolor{white}{49.88}
    \\

    & R1-14B &     
    \cellcolor{mygreen2}57.26&
     \cellcolor{mygreen4}\textcolor{white}{59.63}&
    \cellcolor{mygreen2}57.20&
    \cellcolor{mygreen3}\textcolor{white}{59.14}&

    \cellcolor{mygreen3}\textcolor{white}{58.11}&	
    \cellcolor{mygreen4}\textcolor{white}{59.74}&
        \cellcolor{mygreen1}56.63&	
    \cellcolor{mygreen4}\textcolor{white}{59.44}
    \\

    & QwQ-32B & 
    \cellcolor{mygreen2}69.99&
     \cellcolor{mygreen3}\textcolor{white}{71.75}&
    \cellcolor{mygreen2}69.15&	
    \cellcolor{mygreen3}\textcolor{white}{70.96}&

    \cellcolor{mygreen1}69.71&	
    \cellcolor{mygreen3}\textcolor{white}{71.02}&
    \cellcolor{mygreen2}71.44&	
    \cellcolor{mygreen4}\textcolor{white}{72.00}
    \\
   
    \bottomrule
    \end{tabular}
    \label{tab:random_Pass@1}
\end{table}

To better evaluate the effectiveness of the routing mechanisms introduced in Section~\ref{sec:method}, we use \textit{Random} routing as an extra baseline.
In Table~\ref{tab:random_Pass@1}, we present the Pass@1 results across four routers on four benchmarks.
It is evident that Top-$4$ outperforms Top-$2$ under the \textit{Random} setting, as discussed in Section~\ref{sec:eval_reasoning}. 
To our surprise, when comparing the results across different Routers, we find that \textit{Random} even outperforms \textit{Hybrid}. 
For example, the average Pass@1 of QwQ-32B reaches 71.75 under the Top-$4$ setting, which surpasses \textit{Hybrid} by 0.73.
This is because in the \textit{Random}, the proportion of dissimilar summaries coming from different peers in the model's sampling paths is higher than that of similar summaries.
In contrast, similar and dissimilar summaries each account for half in \textit{Hybrid}.
As a result,the \textit{Random} can introduce more diverse or complementary perspectives, providing the model with more useful information.
Similarly, under the same setting of Top-$4$, the average Pass@1 of QwQ-32B for \textit{Dispersed} reaches 72, which surpasses the \textit{Random} by 0.25, as the \textit{Dispersed} can introduce more diverse or complementary perspectives than \textit{Random}.

\subsection{Cons@N Results}\label{app:cons@N}
To provide a comprehensive evaluation, we also report the Cons@N results in Table~\ref{tab:main_Cons@N}.
We combine the AIME 2024~\cite{aime2024}, AIME 2025~\cite{aime2025}, and AIMO 2025~\cite{ai-mathematical-olympiad-progress-prize-2} benchmarks into a single category, denoted as \textit{Math Comb.}, due to the small size of the individual math benchmarks.
In general, the results under the selected two routing strategies consistently surpass the two baselines by a large margin.
Specifically, on GPQA~\cite{rein2024gpqa}, the Top-$4$ \textit{Hybrid} setting achieves a 10.10 improvement over the Majority Voting~\cite{wang2022self} baseline using DeepSeek-R1-Distill-Qwen-7B~\cite{guo2025deepseek}.
On the average of the three math benchmarks, the Top-$2$ \textit{Hybrid} setting outperforms the baseline by 4.29 on QwQ-32B~\cite{qwq32b}.
Moreover, we observe that the improvement on GPQA~\cite{rein2024gpqa} is more pronounced than on math tasks.
Since GPQA~\cite{rein2024gpqa} requires a combination of factual knowledge and reasoning, the cues learned from peer reasoning paths lead to even greater performance gains.

\begin{table}[H]
    \centering
    \caption{The Cons@N results of Math benchamrks and GPQA Diamond.}
    \begin{tabular}{lcccccccc}
    \toprule
    \multirow{2}{*}{\textbf{Benchmarks}} &\multirow{2}{*}{\textbf{Models}} & \textbf{Majority} & \textbf{Shortest} & \multicolumn{2}{c}{\textbf{Dispersed}} & \multicolumn{2}{c}{\textbf{Hybrid}}\\
    
    \cline{5-8}
    & & \textbf{Vote} & \textbf{Vote} & \textbf{Top-2} & \textbf{Top-4} & \textbf{Top-2} & \textbf{Top-4} \\
    
    \specialrule{0.5pt}{2pt}{2pt}
    \multirow{3}{*}{Math Comb.} 
    & R1-7B   &
    \cellcolor{mygreen1}60.00  &
    \cellcolor{mygreen1}60.00&
    \cellcolor{mygreen3}\textcolor{white}{62.86}  &
    \cellcolor{mygreen2}61.43   &
    \cellcolor{mygreen2}61.43   &
    \cellcolor{mygreen3}\textcolor{white}{62.86}\\
    & R1-14B  &
    \cellcolor{mygreen1}68.57  &
    \cellcolor{mygreen3}\textcolor{white}{70.00}&
    \cellcolor{mygreen1}68.57  &
    \cellcolor{mygreen3}\textcolor{white}{70.00}   &
    \cellcolor{mygreen3}\textcolor{white}{70.00}   &
    \cellcolor{mygreen4}\textcolor{white}{71.43}\\
    & QwQ-32B &
    \cellcolor{mygreen1}78.57  &
    \cellcolor{mygreen2}80.00  &
    \cellcolor{mygreen2}80.00  &
    \cellcolor{mygreen2}80.00   &
    \cellcolor{mygreen4}\textcolor{white}{82.86}   &
    \cellcolor{mygreen3}\textcolor{white}{81.43}\\
    
    \specialrule{0.5pt}{2pt}{2pt}
    \multirow{3}{*}{GPQA Diamond} 
    & R1-7B   &
    \cellcolor{mygreen1}50.00 &
    \cellcolor{mygreen2}51.52 &
    \cellcolor{mygreen4}\textcolor{white}{59.60}  &
    \cellcolor{mygreen3}\textcolor{white}{58.59}  &
    \cellcolor{mygreen2}56.57 &
    \cellcolor{mygreen4}\textcolor{white}{60.10}\\
    & R1-14B  &
    \cellcolor{mygreen1}55.05 &
    \cellcolor{mygreen1}55.05  &
    \cellcolor{mygreen2}{59.60}  &
    \cellcolor{mygreen2}{59.60}  &
    \cellcolor{mygreen3}\textcolor{white}{60.61} &
    \cellcolor{mygreen4}\textcolor{white}{61.11}\\
    & QwQ-32B &
    \cellcolor{mygreen1}60.61 &
    \cellcolor{mygreen1}60.10  &
    \cellcolor{mygreen3}\textcolor{white}{67.68}  &
    \cellcolor{mygreen2}63.64  &
    \cellcolor{mygreen2}63.13 &
    \cellcolor{mygreen4}\textcolor{white}{68.18}\\
   
    \bottomrule
    \end{tabular}
    \label{tab:main_Cons@N}
\end{table}

In order to provide a more detailed analysis of the math tasks, we analyze the AIME 2024 \cite{aime2024}, AIME 2025 \cite{aime2025}, and AIMO 2025 \cite{ai-mathematical-olympiad-progress-prize-2} benchmarks individually.
In Table~\ref{tab:math_Cons@N}, we present the Cons@N results across three math benchmarks.
We observe a significant improvement in both AIME 2024 and AIME 2025 with the two selected routing strategies.
For example, QwQ-32B reaches 86.67 under the Top-4 setting using either the \textit{Dispersed} or \textit{Hybrid} routing strategy on AIME 2024, surpassing the baseline by 3.34.
However, these two routing strategies result in almost no improvement on AIMO 2025.
This is because AIMO has relatively few problems, and the difficulty gap between them is large, making significant improvements hard to achieve.

\begin{table}[H]
    \centering
    \caption{The Cons@32 results on AIME 2024, AIME 2025 and AIMO 2025.}
    \begin{tabular}{lcccccccc}
    \toprule
    \multirow{2}{*}{\textbf{Benchmarks}} &\multirow{2}{*}{\textbf{Models}} & \textbf{Majority} & \textbf{Shortest} & \multicolumn{2}{c}{\textbf{Dispersed}} & \multicolumn{2}{c}{\textbf{Hybrid}}\\
    
    \cline{5-8}
    & & \textbf{Vote} & \textbf{Vote} & \textbf{Top-2} & \textbf{Top-4} & \textbf{Top-2} & \textbf{Top-4} \\
    
    \specialrule{0.5pt}{2pt}{2pt}
    \multirow{3}{*}{AIME 2024} 
    & R1-7B   & 73.33 &73.33& 76.67 & 73.33  & 80.00 & 80.00 \\
    & R1-14B  & 80.00  &80.00   & 80.00    & 83.33  & 80.00 & 80.00 \\
    & QwQ-32B & 83.33 &83.33 & 86.67 & 86.67  & 90.00 & 86.67\\

    \specialrule{0.5pt}{2pt}{2pt}
    \multirow{3}{*}{AIMO 2025} 
    & R1-7B   & 50.00  &50.00 & 50.00 & 50.00 & 50.00  &50.00 \\
    & R1-14B  & 60.00  &60.00& 60.00 & 60.00 & 60.00  &60.00\\
    & QwQ-32B & 70.00  &80.00 & 70.00 & 70.00 & 70.00  &70.00\\
    
    \specialrule{0.5pt}{2pt}{2pt}
    \multirow{3}{*}{AIME 2025} 
    & R1-7B   & 50.00    &50.00 & 53.33    & 53.33    & 46.67    & 50.00 \\
    & R1-14B  & 60.00    &70.00 & 60.00    & 60.00    & 63.33    & 66.67 \\
    & QwQ-32B & 76.67    &70.00 & 76.67    & 76.67    & 80.00    & 80.00\\
    \bottomrule
    \end{tabular}
    \label{tab:math_Cons@N}
\end{table}

\subsection{Results of Number of Tokens and ``Aha'' moments}\label{app:efficiency}
\begin{figure}[H]
  \centering \vspace{-3mm}
  \begin{subfigure}[b]{0.49\textwidth}
        \centering
        \includegraphics[width=\linewidth]{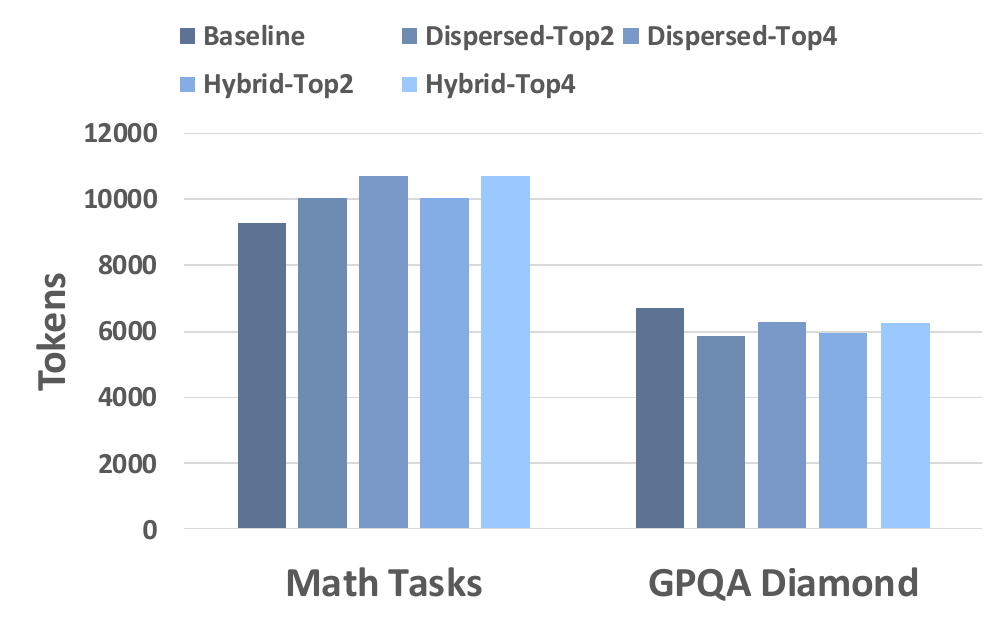}
        \caption{The number of tokens}
        \label{fig:Tokens_7B}
    \end{subfigure}
  \begin{subfigure}[b]{0.49\textwidth}
        \centering
        \includegraphics[width=\linewidth]{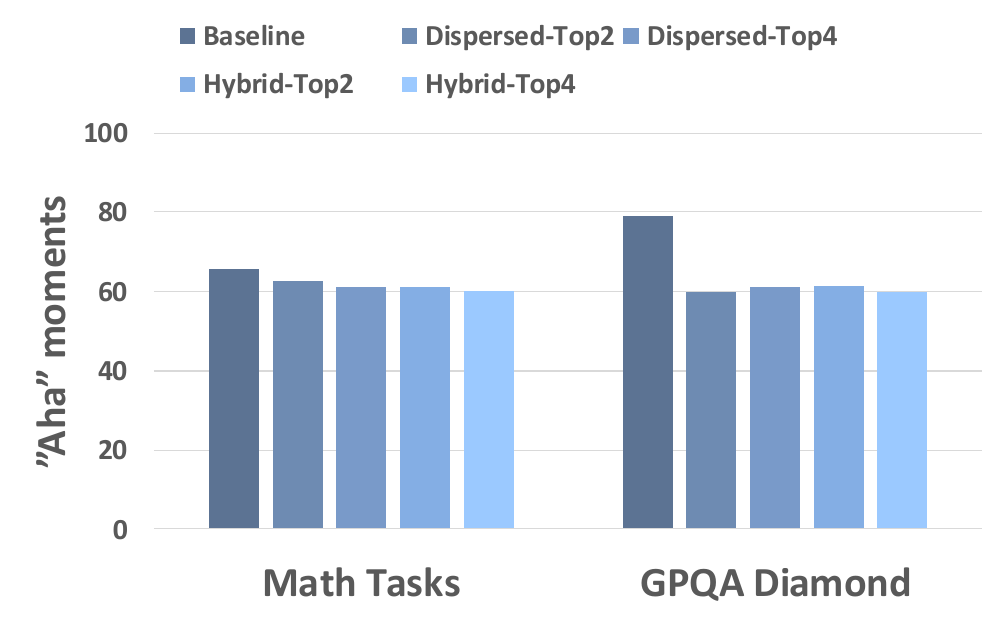}
        \caption{The number of ``Aha'' moments}
        \label{fig:Aha_7B}
    \end{subfigure}
    \caption{We illustrate the average number of tokens and ``Aha'' moments on DeepSeek-R1-Distill-Qwen-7B. Our method produces a comparable number of tokens to the baseline, while yielding fewer ``Aha'' moments.}
    \label{fig:7B_eficiency} \vspace{-3mm}
\end{figure}

\begin{figure}[H]
  \centering \vspace{-3mm}
  \begin{subfigure}[b]{0.49\textwidth}
        \centering
        \includegraphics[width=\linewidth]{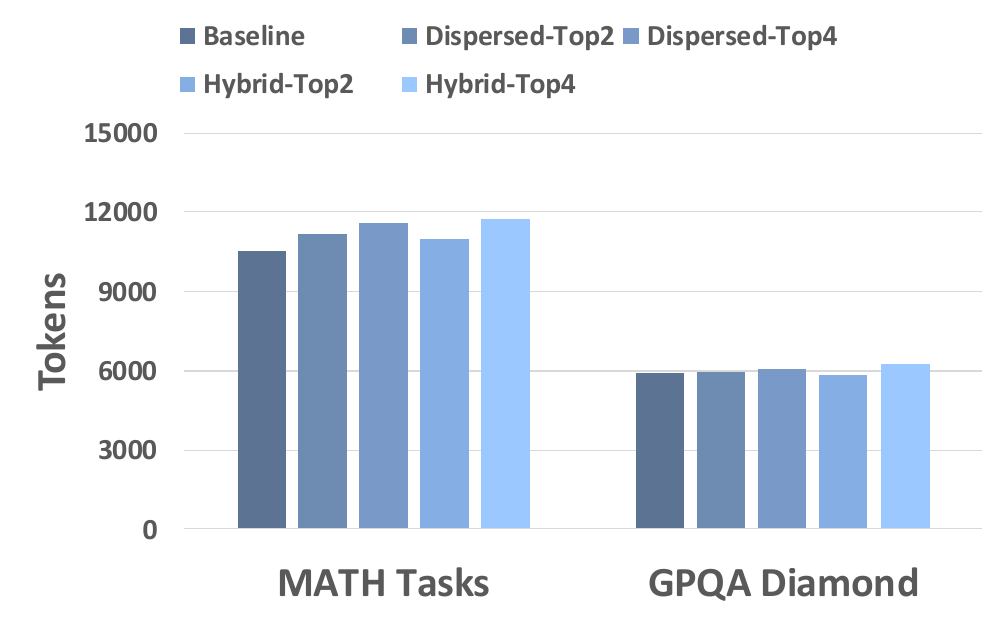}
        \caption{The number of tokens}
        \label{fig:Tokens_14B}
    \end{subfigure}
  \begin{subfigure}[b]{0.49\textwidth}
        \centering
        \includegraphics[width=\linewidth]{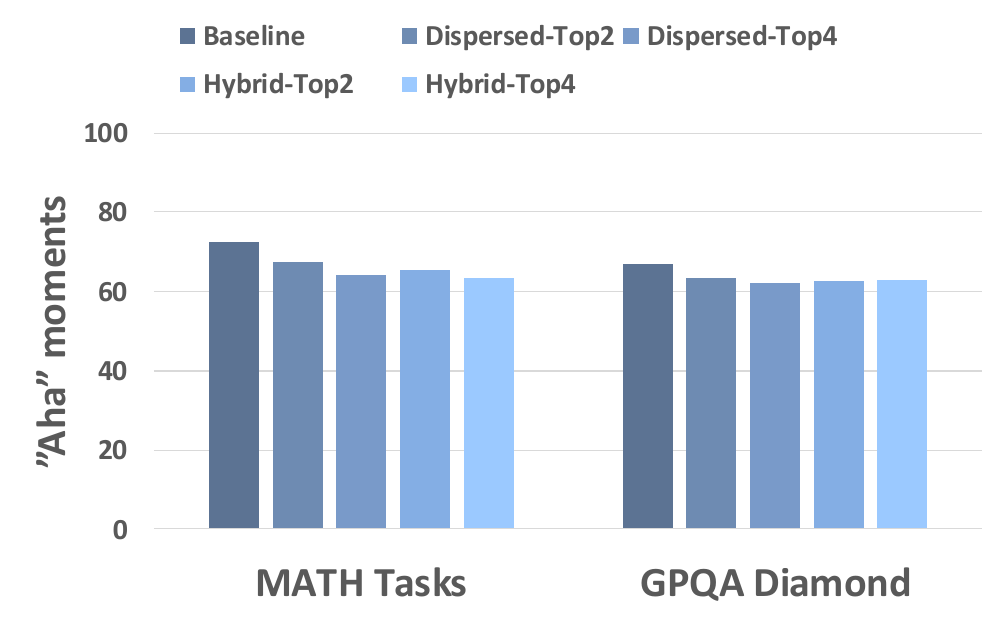}
        \caption{The number of ``Aha'' moments}
        \label{fig:Aha_14B}
    \end{subfigure}
    \caption{We illustrate the average number of tokens and ``Aha'' moments on DeepSeek-R1-Distill-Qwen-14B. Our method produces a comparable number of tokens to the baseline, while yielding fewer ``Aha'' moments.}
    \label{fig:14B_eficiency} \vspace{-3mm}
\end{figure}

We provide more details for the results of LeaP on the DeepSeek-R1-Distill-Qwen-7B and DeepSeek-R1-Distill-Qwen-14B models compared to the baseline in Figures~\ref{fig:7B_eficiency} and~\ref{fig:14B_eficiency}.
We observe that LeaP does not generate significantly more tokens than the baseline for math-related problems.
On the contrary, in the GPQA Diamond, the average length of generated responses is shorter than that of the baseline.
Furthermore, the number of ``aha'' moments for LeaP is generally fewer than that of the baseline in math-related problems and in the GPQA diamond.
This trend is consistent with our analysis on QwQ-32B in Section~\ref{sec:eval_results}.

\section{Case Study}\label{app:case_sutdy}

\subsection{Case Study on Comparing LeaP and Baseline with Bad Beginning}\label{app:case_study_bad_begin}

\begin{figure}[H]
\centering
\begin{prompt}{Case Study on LeaP with Bad Beginning}
\textcolor{blue}{Problem:}

Let $A$, $B$, $C$, and $D$ be point on the hyperbola $\frac{x^2}{20}- \frac{y^2}{24} = 1$ such that $ABCD$ is a rhombus whose diagonals intersect at the origin. Find the greatest real number that is less than $BD^2$ for all such rhombi.
\\
\\
\textcolor{blue}{Model Response:}
\textcolor{darkred}{Okay, so I have this problem where there are four points A, B, C, D on the hyperbola given by the equation \( \frac{x^2}{20} - \frac{y^2}{24} = 1 \).
...
So, we need to find the supremum (the least upper bound) of \( BD^2 = 4(r^2 + s^2) \), such that the points satisfy the hyperbola equation and the condition \( pr + qs = 0 \). So, let me formalize this. Given: 1. \( \frac{p^2}{20} - \frac{q^2}{24} = 1 \). 2. \( \frac{r^2}{20} - \frac{s^2}{24} = 1 \). 3. \( pr + qs = 0 \). We need to find the supremum of \( 4(r^2 + s^2) \). \textbf{(Bad Beginning)}}

...

\textcolor{darkred}{<summarize> In conclusion, the current key takeaways and results are: ... Using hyperbola parametrization and vector properties, I derived an expression for BD² in terms of a parameter b, which is tanh u. The expression for BD² is \( \frac{80 + 96 b^2}{1 - b^2} \) with the constraint \( |b| \in (\frac{5}{6}, 1) \). By analyzing the derivative of this function, I found that it's increasing in the interval, implying the minimum BD² occurs as b approaches \( \frac{5}{6} \). Therefore, the minimal BD² is achieved in the limit as b approaches \( \frac{5}{6} \). </summarize> \textbf{(Incorrect Summary)}}

<peer\_summaries>

Peer 1: "To summarize, based on my previous reasoning, I have currently found that $BD^2$ can be expressed as $80 + (22v)/3$, where $v = s^2$. The variable v is constrained by the equation $u = [600(v - 24)] / (11v - 600)$, where $u = q^2$. The variable v must be either less than 24 or greater than approximately 54.545...

Peer 2: "In short, I've currently concluded that the expression for \( BD^2 \) in terms of \( u = p^2 \) is \( 480 \times \frac{11u - 120}{11u - 720} \). I also noticed that as \( u \) approaches \( \frac{720}{11} \) from above, \( BD^2 \) tends to infinity, which seems contradictory because points on a hyperbola shouldn't allow for infinitely long diagonals...

Peer 3: "To summarize, my recent findings are: I attempted to express \( p \) and \( s \) in terms of \( q \) and \( r \) using the condition \( pr + qs = 0 \), leading to \( p = k q \) and \( s = -k r \)... 

\textcolor{darkgreen}{Peer 4: "In conclusion, the current insights and results I've gathered are: Expressed BD² in terms of t (where t = q²), transformed the equation into a function f(t) = 80 + [4400(t + 24)] / (11t - 600), and simplified it to f(t) = 480 + (345600/11)/(t - 600/11). The domain of t is t > 600/11. So, as t approaches infinity, f(t) approaches 480, and as t approaches 600/11 from the right, f(t) approaches infinity. Therefore, BD² can be made arbitrarily large, but the problem asks for the greatest real number less than BD² for all such rhombi. So, that number is 480." \textbf{(Correct Summary from Peers)}} </peer\_summaries>

...

 From the comments, it appears that some peers have approached the problem differently and arrived at different expressions for BD².\textcolor{darkgreen}{ Notably, Peer 4 suggests that BD² approaches 480 as t approaches infinity, and that BD² can be made arbitrarily large, but the greatest number less than BD² for all rhombi is 480. \textbf{(Self-Verification)}} Wait, that seems conflicting with my earlier conclusion where BD² approaches infinity as b approaches 1. So, perhaps I made a mistake in my parametrization. Wait, let me check Peer 4's approach. Peer 4 says: Expressed BD² in terms of t (where t = q²)
 
 ...
 
 Therefore, the infimum of BD² is 480, so the greatest real number less than BD² for all rhombi is 480. But in my own reasoning, I thought BD² can be made as large as desired, which would suggest that the infimum is 480, and the supremum is infinity. So, the greatest real number less than BD² for all rhombi is 480...
 
 Therefore, the answer is 480. But wait, in my earlier parametrization, I had BD² tending to infinity as b approaches 1, but here, it's showing that BD² can be made as small as 480 and as large as infinity. \textcolor{darkgreen}{Therefore, the greatest real number less than BD² for all rhombi is 480. So, the answer is 480. **Final Answer** \boxed{480}}
\end{prompt}
\caption{Case Study on LeaP with bad beginning.}
\label{fig:comparison_leap}
\end{figure}

\begin{figure}[H]
\centering
\begin{prompt}{Case Study on Baseline with Bad Beginning}
\textcolor{blue}{Problem:}

Let $A$, $B$, $C$, and $D$ be point on the hyperbola $\frac{x^2}{20}- \frac{y^2}{24} = 1$ such that $ABCD$ is a rhombus whose diagonals intersect at the origin. Find the greatest real number that is less than $BD^2$ for all such rhombi.
\\
\\
\textcolor{blue}{Model Response:}

\textcolor{darkred}{Okay, so I have this problem where there are four points A, B, C, D on the hyperbola given by the equation \( \frac{x^2}{20} - \frac{y^2}{24} = 1 \).
...
So, we need to find the supremum (the least upper bound) of \( BD^2 = 4(r^2 + s^2) \), such that the points satisfy the hyperbola equation and the condition \( pr + qs = 0 \). So, let me formalize this. Given: 1. \( \frac{p^2}{20} - \frac{q^2}{24} = 1 \). 2. \( \frac{r^2}{20} - \frac{s^2}{24} = 1 \). 3. \( pr + qs = 0 \). We need to find the supremum of \( 4(r^2 + s^2) \). \textbf{(Bad Beginning)}}

...

\textcolor{darkred}{Therefore, f(t) decreases from +infinity to 80 as t increases from 5/6 to infinity. Therefore, the infimum of BD² is 80, but it's never actually reached; it's a limit as t approaches infinity. Therefore, BD² can be made arbitrarily close to 80, but never less than 80. Therefore, the greatest real number less than BD² for all such rhombi is 80. Hence, the answer is \boxed{80}. **Final Answer** \boxed{80}}
 
\end{prompt}
\caption{Case Study on baseline with bad beginning.}
\label{fig:comparison_baseline}
\end{figure}

\subsection{Case Study on Failure Examples}\label{app:case_sutdy_failure}

We provide some case studies to better illustrate our approach.
In Figure~\ref{fig:case_study_summary}, we present a failure case when applying LeaP to DeepSeek-R1-Distill-Qwen-7B.
In this example, the model fails to clearly summarize its own reasoning, which results in the loss of partial reasoning information during summarization.
This issue primarily relates to model size.
Specifically, when the model is relatively small, incomplete summarization tends to occur more frequently.

In the second case, we demonstrate a situation where the model refuses to engage in communication.
As shown in Figure~\ref{fig:case_study_LeaP}, the model assumes it is alone and therefore ignores the perspectives of other peers, continuing its own reasoning trajectory.
This case is observed in QwQ-32B but not in the distilled model from R1.
We hypothesize that this difference may be related to the training method: QwQ-32B is trained via reinforcement learning (RL), which may lead the model to exhibit higher confidence.

\begin{figure}[H]
\centering
\begin{prompt}{Case Study on Summarization Failure}
\textcolor{blue}{Problem:}

Every morning Aya goes for a $9$-kilometer-long walk and stops at a coffee shop afterwards. When she walks at a constant speed of $s$ kilometers per hour, the walk takes her 4 hours, including $t$ minutes spent in the coffee shop. When she walks $s+2$ kilometers per hour, the walk takes her 2 hours and 24 minutes, including $t$ minutes spent in the coffee shop. Suppose Aya walks at $s+\frac{1}{2}$ kilometers per hour. Find the number of minutes the walk takes her, including the $t$ minutes spent in the coffee shop.
\\
\\

\textcolor{blue}{Model Response:}

...

Wait, wait, from $x^{253} -34x +33=0$, $x= k^8$, so $x=1$ is a root, so we can factor it as $(x -1)(something)=0$. But since $x=1$ is a root, and we're interested in the other root, perhaps it's a very small x, but I'm not sure.
Wait, perhaps I can write the equation as $x^{253}=34x -33$.
Hold on, I should summarize the key points briefly to ensure I'm on the right track.
\textcolor{darkred}{<summarize> To summarize, my recent findings are: The problem involves two circles of radius 34 and 2024 arranged in a triangle ABC, with each circle tangent to two sides and the previous circle. The inradius $R$ of ABC can be expressed as $R=34 \cdot \frac{1 -k^8}{1 -k}$ for the 34-radius circles, and $R=\frac{1 -k^{2024}}{1 -k}$ for the 1-radius circles. Setting these equal gives the equation $34 \cdot (1 -k^8) = 1 -k^{2024}$, which simplifies to $k^{2024} - 34k^8 + 33 = 0$. Letting $x = k^8$, this becomes $x^{253} - 34x + 33 = 0$, which factors as $(x - 1)(x^{252} + x^{251} + \cdots + x + 1) - 34x + 33 = 0$. Since $x = 1$ is a root, but we need another real root where $x < 1$. Approximating, $x \approx 0.970588$, leading to $k \approx 0.9963$. Then, $R \approx$ ... (\textbf{truncated at 256 tokens}) </summarize>}

...
\end{prompt}
\caption{The case study about the reasoning model fail to summarize concisely.}
\label{fig:case_study_summary}
\end{figure}

\begin{figure}[H]
\centering
\begin{prompt}{Case Study on Reflection Failure}
\textcolor{blue}{Problem:}

Torus $T$ is the surface produced by revolving a circle with radius $3$ around an axis in the plane of the circle that is a distance $6$ from the center of the circle (so like a donut). Let $S$ be a sphere with a radius $11$. When $T$ rests on the inside of $S$, it is internally tangent to $S$ along a circle with radius $r_i$, and when $T$ rests on the outside of $S$, it is externally tangent to $S$ along a circle with radius $r_o$. The difference $r_i-r_o$ can be written as $\tfrac{m}{n}$, where $m$ and $n$ are relatively prime positive integers. Find $m+n$. [asy] unitsize(0.3 inch); draw(ellipse((0,0), 3, 1.75)); draw((-1.2,0.1)..(-0.8,-0.03)..(-0.4,-0.11)..(0,-0.15)..(0.4,-0.11)..(0.8,-0.03)..(1.2,0.1)); draw((-1,0.04)..(-0.5,0.12)..(0,0.16)..(0.5,0.12)..(1,0.04)); draw((0,2.4)--(0,-0.15)); draw((0,-0.15)--(0,-1.75), dashed); draw((0,-1.75)--(0,-2.25)); draw(ellipse((2,0), 1, 0.9)); draw((2.03,-0.02)--(2.9,-0.4)); [/asy]
\\
\\
\textcolor{blue}{Model Response:}

...

<peer\_summaries>

...

Peer 4: "In conclusion, the current insights and results I've gathered are: By setting up the equations for the sphere and the torus, and analyzing the gradients for tangency, I derived that at points of tangency, the z-component gives $k = 2$ (assuming $z \neq 0$).
This led to an equation relating the radial distance $\rho = sqrt(x^2 + y^2)$ to the major radius $R =6$. Solving this gives a specific $\rho$ value, which when substituted back into the equations, yields a valid solution for $z$.
This allows computing the radii $r_i$ and $r_o$ for internal and external tangency, leading to the desired difference. However, I must ensure that the configuration (internal vs. external) corresponds correctly to the direction of the normals." </peer\_summaries>
Hmm, it seems that my peers have given me some comments, so let me check if anyone's conclusions are different from mine before I continue my own reasoning.
\textcolor{darkred}{But since I'm alone, I'll proceed. Let me try to proceed with the standard equation of the torus. Let me define the torus such that its $axis$ is the $x-axis$.}
The generating circle has a major radius $R = 6$, so its center is at (6, 0, 0), and minor radius $r = 3$. The equation of the torus is: $( sqrt(y^2 + z^2) )^2 + (x - 6)^2 = 3^2$ ? Wait, no, perhaps I need to think differently. Let me recall that the standard equation for a torus rotated around the $x-axis$ with major radius R and minor radius r is: $( sqrt( (sqrt(y^2 + z^2) - R )^2 + x^2 ) )$? Wait, perhaps I should use a different approach. 

...
\end{prompt}
\caption{The case study that the QwQ-32B reject to understand the summaries from peers.}
\label{fig:case_study_LeaP}
\end{figure}

\subsection{Case Study on Success Examples}\label{app:case_study_success}

In the third case, we present a scenario where the model ultimately reaches the correct answer through interaction with its peers. As shown in Figure~\ref{fig:case_study_communication}, the model initially fails to fully summarize its own reasoning process, leading to the loss of partial reasoning information during summarization. However, during the subsequent interaction with peers, the model receives a clearer and more explicit reasoning path provided by them. By verifying the summaries generated by other peers, the model gradually identifies the correct solution to the problem.

In the fourth case, we present a scenario where the model initially generates an incorrect answer during the reasoning process. Meanwhile, other peers also produce flawed summaries, though some provide correct ones.
Through interaction and mutual verification, the model eventually identifies the correct reasoning path. As shown in Figure~\ref{fig:case_study_communication_two}, the model first summarizes its own reasoning and arrives at the incorrect answer of 389. Peer 1 and Peer 2 provide incorrect summaries, while only Peer 3 offers a reasonable one. By verifying the summaries from all peers, the model determines that the summaries from Peer 1 and Peer 2 are inconsistent. In contrast, the summary from Peer 3 is considered reliable. Based on this verification process, the model gradually uncovers the correct solution to the problem.

\begin{figure}[H]
\centering
\begin{prompt}{Case Study on LeaP Success}
\textcolor{blue}{Problem:}

"Each vertex of a regular octagon is independently colored either red or blue with equal probability. The probability that the octagon can then be rotated so that all of the blue vertices end up at positions where there were originally red vertices is $\tfrac{m}{n}$, where $m$ and $n$ are relatively prime positive integers. What is $m+n$?"
\\
\\
\textcolor{blue}{Model Response:}

...

<peer\_summaries>

...

<summarize> In conclusion, the current insights and results I've gathered are as follows: - **Good Subsets**: Subsets with 0 to 3 elements are all good, totaling \(1 + 8 + 28 + 56 = 93\) subsets. Subsets of size 4 may include some bad subsets, while subsets of size 5–8 are mostly bad, though not all.
- **Bad Subsets of Size 4**: Examples like {0,1,3,4} cannot be rotated to be disjoint, indicating that some size-4 subsets are bad. Calculating their exact count requires identifying patterns or symmetries that make these subsets resistant to rotation. - **Symmetry and Complementarity**: 
The problem's symmetry suggests that the number of bad subsets of size \(k\) equals those of size \(8 - k\).
However, since subsets of size 3 (complement of 5) are all good, subsets of size 5 must also be good, contradicting earlier assumptions.\textcolor{darkred}{ This implies a need to reevaluate. - **Reevaluation**: Perhaps subsets of size 5 are actually good, and the real bad subsets are only those of size 7, 8, and some specific size 4 subsets. This would mean the total bad subsets are 1 (... </summarize> (\textbf{truncated at 256 tokens})}
 <peer\_summaries> Peer 1: "..."Peer 2:"..."Peer 3: "..."\textcolor{darkgreen}{Peer 4:"To summarize, my recent findings are as follows: - **Subsets of Size 4**: - **Rotation by 1 or 7**: Only two valid subsets (checkerboard patterns {0,2,4,6} and {1,3,5,7}).
 - **Rotation by 2 or 6**: Requires selecting opposite pairs in each of the two 4-cycles formed by the rotation. This yields $2 \times 2 =4$ subsets.
 - **Rotation by 4**: Selecting one vertex from each of the four diametrical pairs, resulting in $2^4 =16$ subsets. - **Overlap Considerations**: No overlaps between these categories, as the subsets from different rotations are structurally distinct. - **Total Valid Size 4 Subsets**: 2 +4 +16 =22.
 - **Total Favorable Subsets**: - Size 0:1 - Size 1:8 - Size 2:28 - Size 3:56 - Size 4:22 - **Total**:1+8+28+56+22 =115 - **Probability Calculation**: - Total subsets:..."</peer\_summaries>}
 Hmm, it seems that my peers have given me some comments, so let me check if anyone's conclusions are different from mine before I continue my own reasoning.
Peer 1 is talking about independent sets on a cycle with distance constraints. Peer 2 suggests that size4 subsets can be counted by analyzing different rotation types, leading to 22 valid subsets.
 Peer 3 mentions inclusion-exclusion over rotation classes.  \textcolor{darkgreen}{Peer 4 calculates total favorable subsets as 115, leading to probability 115/256. Wait, Peer4's calculation: Size0:1 Size1:8 Size2:28 Size3:56 Size4:22 (only those that are good) Total: 1+8+28+56+22 =115 Thus probability 115/256, which simplifies to 115 and 256 are coprime?
 256 is $2^8$. 115 is $5 \times 23$, so yes. Thus m+n=115+256=371.}
 But I need to verify if this is correct. According to Peer4's reasoning: For size4 subsets, the valid (good) subsets are 22. The bad subsets of size4 are 70-22=48. Therefore, total bad subsets are: Size5:56 Size6:28 Size7:8 Size8:1 Size4 bad:48 Total bad:56+28+8+1+48=141 Thus total good subsets:256-141=115 Therefore, probability 115/256 Thus the answer is 115+256=371. But I need to verify if Peer4's count of size4 subsets is correct. Peer4 says
 
 ...

\textcolor{darkgreen}{
 Yes. Thus, these subsets are good. The checkerboard patterns are also good. Therefore, the count of22 valid subsets of size4 is correct. Thus, the total good subsets are115, leading to the probability of115/256. Thus, the final answer is115 +256= 371.(\textbf{ {**Final Answer** \boxed{371} })}}
\end{prompt}
\caption{The case study that the model learn from peers successfully, which lead to a correct final answer.}
\label{fig:case_study_communication}
\end{figure}

\begin{figure}[H]
\centering
\begin{prompt}{Case Study on LeaP Success}
\textcolor{blue}{Problem:}

"Each vertex of a regular octagon is independently colored either red or blue with equal probability. The probability that the octagon can then be rotated so that all of the blue vertices end up at positions where there were originally red vertices is $\tfrac{m}{n}$, where $m$ and $n$ are relatively prime positive integers. What is $m+n$?"
\\
\\
\textcolor{blue}{Model Response:}

...

 <summarize> In conclusion, the current insights and results I've gathered are as follows: - The problem involves two configurations of circles arranged in triangle ABC: eight circles of radius 34 and 2024 circles of radius 1,
 
 ...
 
 \textcolor{darkred}{- After extensive exploration and considering peer insights, the most consistent approach involves setting up equations based on the path lengths required by both configurations and solving for the inradius using proportional reasoning, leading to an inradius of 
  $( \frac{384}{5} )$,(\textbf{ resulting in $( m + n = 384 + 5 = 389  ...$ })}</summarize> <peer\_summaries>\textcolor{darkred}{ Peer 1: "To summarize, based on my previous reasoning, I have currently found }
  
 ...
 
  \textcolor{darkred}{Thus, GCD is 1, making \( \frac{68,800}{2007} \) already in simplest terms. Adding numerator and denominator gives \( 68,800 + 2007 = 70,807 \). However, this contrad..." Peer 2: "In short, my current conclusions are that the problem involves arranging circles in geometric progression along an angle bisector of triangle ABC's vertex B. }
  
  ...
  
 \textcolor{darkred}{ The relationship \( 2024 = 8 \times 253 \) hints that \( k^{2016} = 34 \) simplifies to \( (k^8)^{252} = 34 \), implying \( k^8 = \sqrt[252]{34} \). This still leaves the inradius..."}
  \textcolor{darkgreen}{ Peer 3: "In short, I've currently concluded that by modeling the problem using a geometric progression of radii along the angle bisector at vertex B, and setting up equations for both configurations (8 circles of radius 34 and 2024 circles of radius 1), I derived that the sine of half the angle \( \theta/2 \) is \( \frac{11}{1190} \). Substituting back into the inradius formula yields \( R = \frac{192}{5} \), which simplifies to \( \frac{192}{5} \). Since 192 and 5 are coprime, the final answer is \( m + n = 197 \). This result aligns with the calculations when assuming the incenter is positioned such that the distance from vertex B to the incenter along the angle bisector equals the distance to the last circle plus its radius." }Peer 4: "In conclusion, the current key takeaways and results are as follows: By modeling the problem using geometric progressions for the radii of the circles arranged along the angle at vertex B, 
  
  ...
  
  the expression for \( R \) simplifies to \( R = \frac{2 \times 34^{253/252}}{34^{1/2016} + 1} \). Recognizing that \( 34^{253/252} = 34^{1 + 1/252} \), this becomes R =..." </peer\_summaries> 
  
  ...
  
  \textcolor{darkgreen}{ Peer 3 suggests an inradius of 192/5, so 192+5=197.} \textcolor{darkred}{ Peer 1 and 2 are getting large fractions. Peer 4's answer is unclear.}
  Given that in the Asymptote example, the inradius was about 0.45 for small circles, and the problem has 2024 circles of radius 1, 

  ...

  \textcolor{darkgreen}{
  which is not a fraction. Alternatively, the answer is 34/253 +1/2024? No. Alternatively, the problem is a trick question where the inradius is simply the sum of the radii multiplied by the number of circles divided by some number. Alternatively, the inradius is the solution to the equation: 8*(2*34) =2024*(2*1) Which is 544 =4048 $\to$ impossible. Thus, the only plausible answer is Peer3's 192/5, which is 38.4.}
  
  ...
  
  \textcolor{darkred}{
  Given that 384/5 is 76.8, and the path length for first case is 544, which is 7*68 + 68=544, and 544/7=77.714. Close to 76.8. Perhaps Peer1's calculation is correct. Alternatively, the inradius is 544/7=77.714= 544/7=77.714285..., which is 544 divided by7 is 77.714. 544 divided by7:7$\times$77=539, so 544/7=77+5/7=77.714285... Thus, as a fraction, 544/7, which reduces to 544/7 (since 544 and7 are coprime). Thus, m+n=544+7=551.}

 …

\textcolor{darkgreen}{
 then r=34 +476 sin$\theta$ To also satisfy the second configuration: r=1 +4046 sin$\theta$ Thus, 34+476 sin$\theta$=1+4046 sin$\theta$ 33= 3570 sin$\theta$ sin$\theta$=33/3570=11/1190$=$0.0092 Thus, sin$\theta$$=$0.0092 Then, r=1 +4046*(11/1190)=1 + (4046$\times$11)/1190 Calculate numerator:4046$\times$11=44506 Divide by1190: 44506/1190 $=$37.4 Thus, r $=$1+37.4=38.4=384/10=192/5=38.4 Ah, so this gives r=192/5=38.4 Thus, m=192, n=5, so m+n=197 This matches Peer3's answer. Therefore, the inradius is 192/5. The steps: Assuming the progression ratio k=1 (even though impossible), then the inradius is derived as above, leading to sin$\theta$=11/1190, and substituting back gives r=192/5. \textbf{Thus, the answer is 192+5=197. **Final Answer** \boxed{197}}})
...
\end{prompt}
\caption{The case study that the model learn from peers successfully, which lead to a correct final answer.}
\label{fig:case_study_communication_two}
\end{figure}

\section{Training Hyperparameters of LeaP-T}\label{app:hyper_leapt}

\begin{table}[H]
\centering
\caption{Hyperparameters for LeaP-T training.}
\label{tab:sft_hyperparameters}
\begin{tabular}{lccc}
\toprule
\textbf{Hyperparameter} & \textbf{1.5B Model} & \textbf{7B Model} & \textbf{14B Model}\\
\midrule
Batch size & 16 & 16 & 16 \\
Learning rate & $1 \times 10^{-5}$ & $1 \times 10^{-5}$ & $1 \times 10^{-5}$\\
Learning rate scheduler & Cosine decay & Cosine decay & Cosine decay\\
Warmup ratio & 0.05 & 0.05 & 0.05 \\
Optimizer & AdamW & AdamW & AdamW\\
Weight decay & $1 \times 10^{-4}$ & $1 \times 10^{-4}$ & $1 \times 10^{-4}$\\
Max sequence length & 16K & 16K & 16K \\
Training epochs & 8.0 & 8.0 & 5.0 \\
Precision & bfloat16 & bfloat16 & bfloat16 \\
\bottomrule
\end{tabular}
\end{table}

\section{Comparison with MoA}\label{app:moa}

\begin{figure}[H]
    \centering
    \includegraphics[width=.5\linewidth]{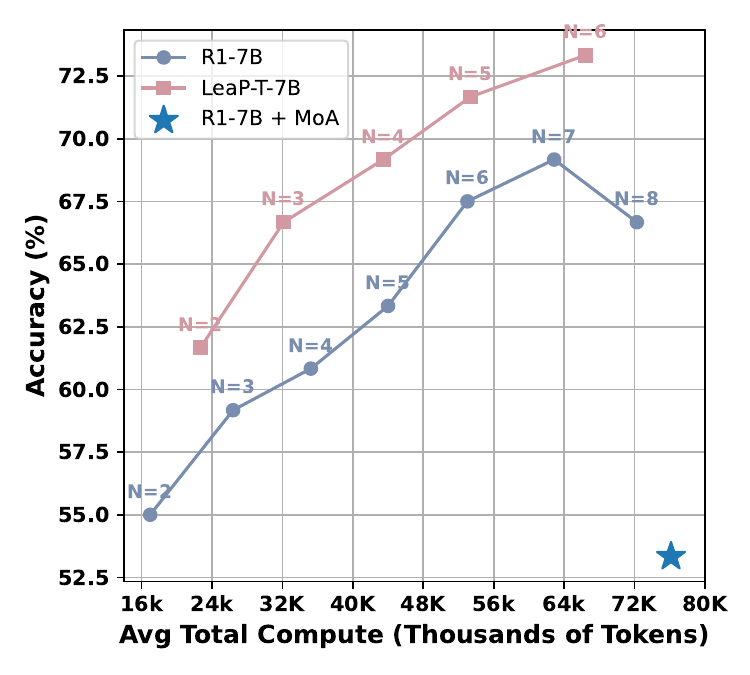}
    \caption{Accuracy vs. Total tokens on AIME 2024.}
    \label{fig:scaling_with_moa}
\end{figure}
We also test MoA~\cite{wang2024mixture} as a baseline under the setting of $4$ Layers and $3$ Agents, with the prompts available in Appendix~\ref{app:prompts}.
As shown in Figure~\ref{fig:scaling_with_moa}, MoA costs approximately 80K tokens per problem, but the results show that it does not transfer well to LRMs.
The reason is that LRMs cannot follow the user’s instructions effectively. We find that even when the correct answer is achieved in the intermediate layers, the model still reaches incorrect conclusions in the final layer.

\section{Other In-depth Analysis}\label{app:other_analysis}
\subsection{Sensitivity Analysis}\label{app:sensitivity}
\subsubsection{On Temperature}\label{app:ablation_temp}

\begin{figure}[H]
    \centering
    \includegraphics[width=.5\linewidth]{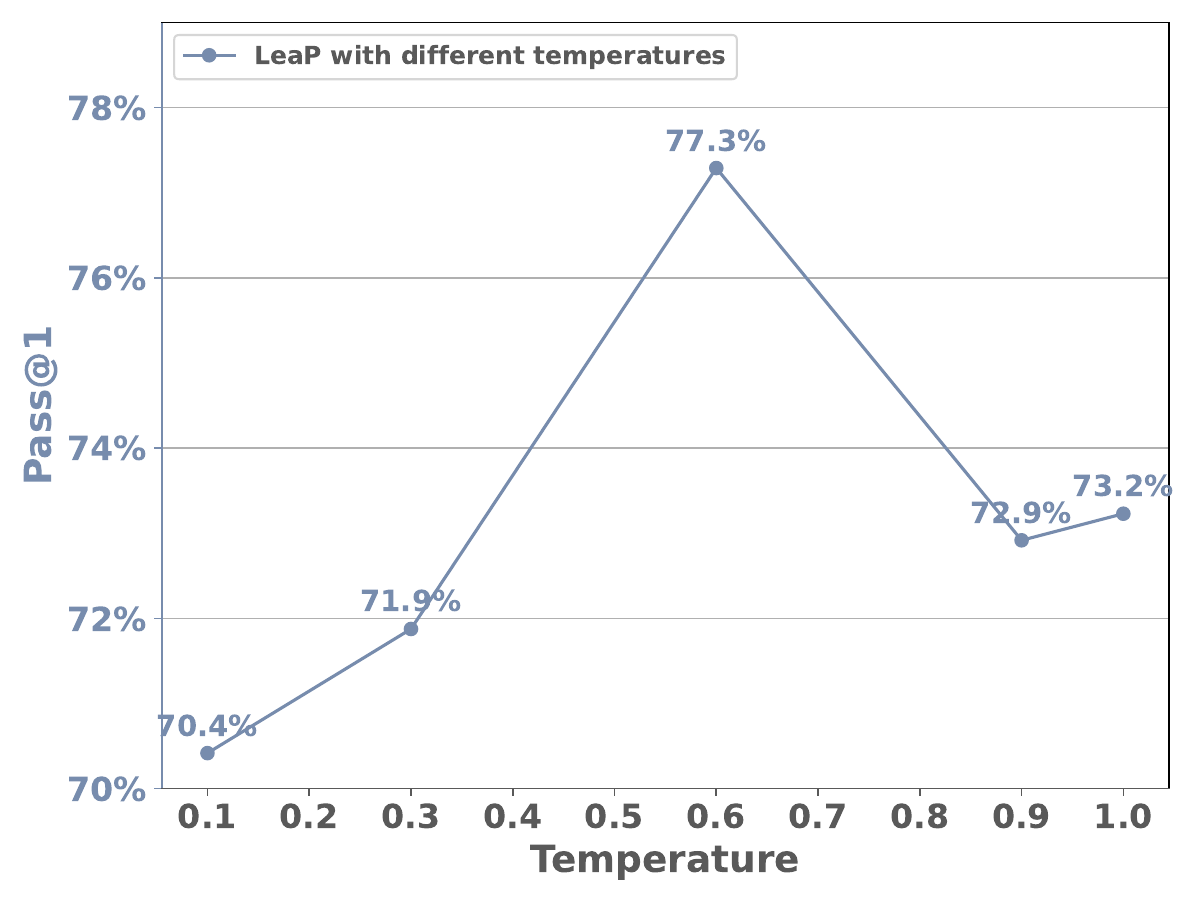}
    \caption{Pass@1 results of various temperatures on AIME 2024.}
    \label{fig:Temperature}
\end{figure}
We conduct an in-depth analysis of the temperature parameter $\tau$ by varying it from 0.1 to 1.0, while keeping all other settings fixed. We use Top-$4$ Dispersed routing and set the communication interval to $T=4$k tokens. As shown in Figure~\ref{fig:Temperature}, when the temperature is low ($\tau \leq 0.3$), the model achieves an average accuracy of around 71, similar to the \textit{Clustered} router baseline. This is expected, as a low temperature reduces output diversity, thereby limiting the benefits of peer communication. On the other hand, when the temperature is too high (e.g., $\tau \geq 0.9$), performance drops noticeably. This is likely due to excessive randomness, which may lead the model away from coherent reasoning or instruction following. Overall, a moderate temperature appears to strike a good balance between diversity and stability.

\subsubsection{On Communication Granularity $T$ for LeaP-T}\label{app:leap_t_granularity}
\begin{figure}[H]
  \centering \vspace{-3mm}
  \begin{subfigure}[b]{0.49\textwidth}
        \centering
    \includegraphics[width=\linewidth]{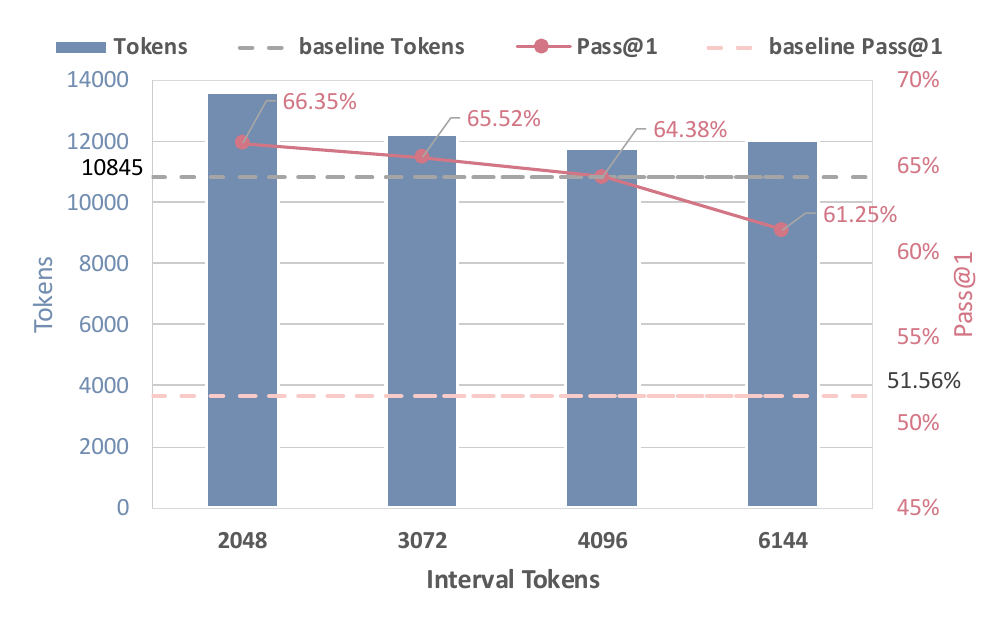}
        \caption{7B}
        \label{fig:Interval_Tokens_7B_sft}
    \end{subfigure}
  \hspace{-2mm}
  \begin{subfigure}[b]{0.49\textwidth}
        \centering
        \includegraphics[width=\linewidth]{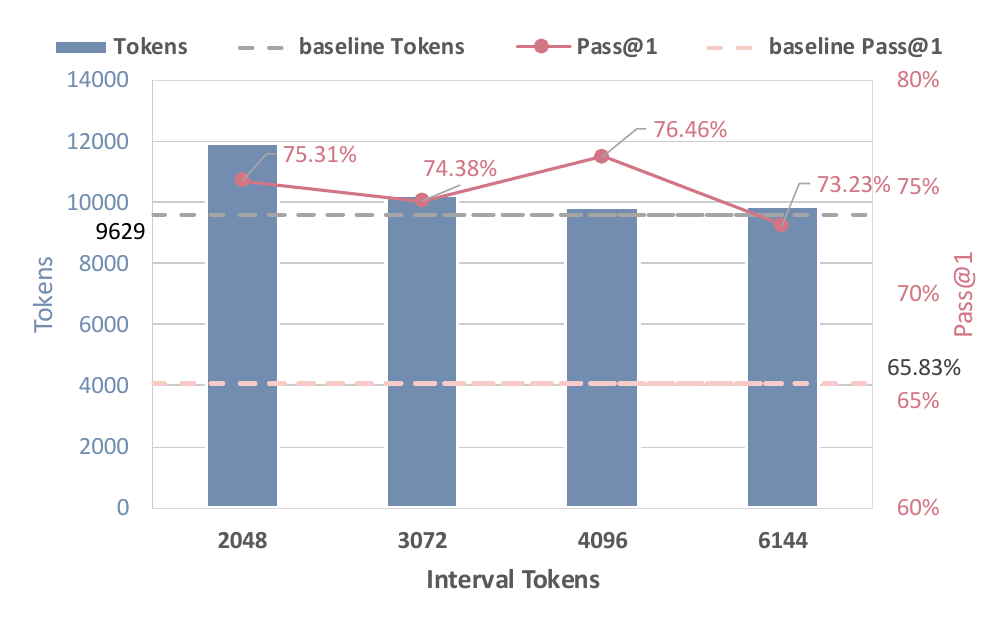}
        \caption{14B}
        \label{fig:Interval_Tokens_14B_sft}
    \end{subfigure}
    \caption{Pass@1 and total tokens on AIME 2024 for LeaP-T-7B and LeaP-T-14B models, evaluated across interval tokens of LeaP from 2048 to 6144.}
    \label{fig:ablation_freguency_sft} \vspace{-3mm}
\end{figure}

To further study the effect of communication granularity, we analyze LeaP-T under different token intervals \( T \). As shown in Figure~\ref{fig:Interval_Tokens_7B_sft}, for the 7B version, the Pass@1 score on AIME 2024 decreases from 66.35 to 61.25 as \( T \) increases, while the number of generated tokens drops accordingly. This trend is consistent with the analysis in Section~\ref{sec:analysis_granularity}: more frequent communication (i.e., smaller \( T \)) improves performance slightly, but increases token consumption due to more frequent summarization and message exchange.
We observe a similar trend for the 14B version in Figure~\ref{fig:Interval_Tokens_14B_sft}.

\subsubsection{On Communication Position for LeaP-T}\label{app:leap_t_position}

\begin{figure}[H]
    \centering
    \includegraphics[width=.5\linewidth]{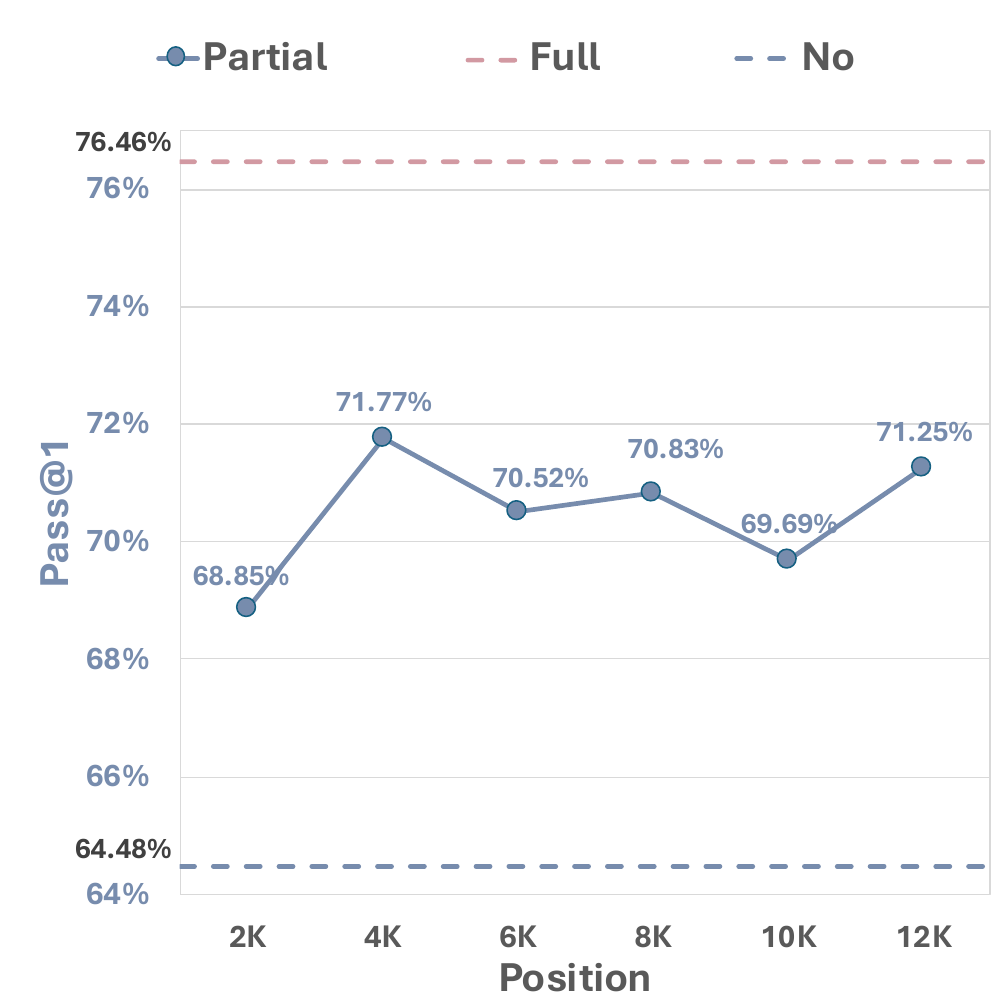}
    \caption{Performance of partial LeaP on the 14B model with LeaP-T, evaluated across the positions of LeaP block ranging from 2K to 12K tokens}
    \label{fig:Position-14B-sft}
\end{figure}

Furthermore, Figure~\ref{fig:Position-14B-sft} analyzes when communication is most effective. The results indicate that performance peaks when the LeaP block is inserted at 4K tokens. Specifically, Pass@1 increases from 68.85 at 2K to 71.77, then declines to 69.69 at later positions. Although a slight recovery to 71.25 is observed, the overall pattern suggests that earlier communication tends to yield better results. These results are consistent with our analysis in Section~\ref{sec:analysis_position}, and highlight the importance of timely information exchange in improving performance.

\subsection{Robustness at Various Difficulty Levels for LeaP-T}\label{app:leap_t_diff}
\label{LeaP-new-Difficulty}

\begin{figure}[H]
    \centering
    \includegraphics[width=.5\linewidth]{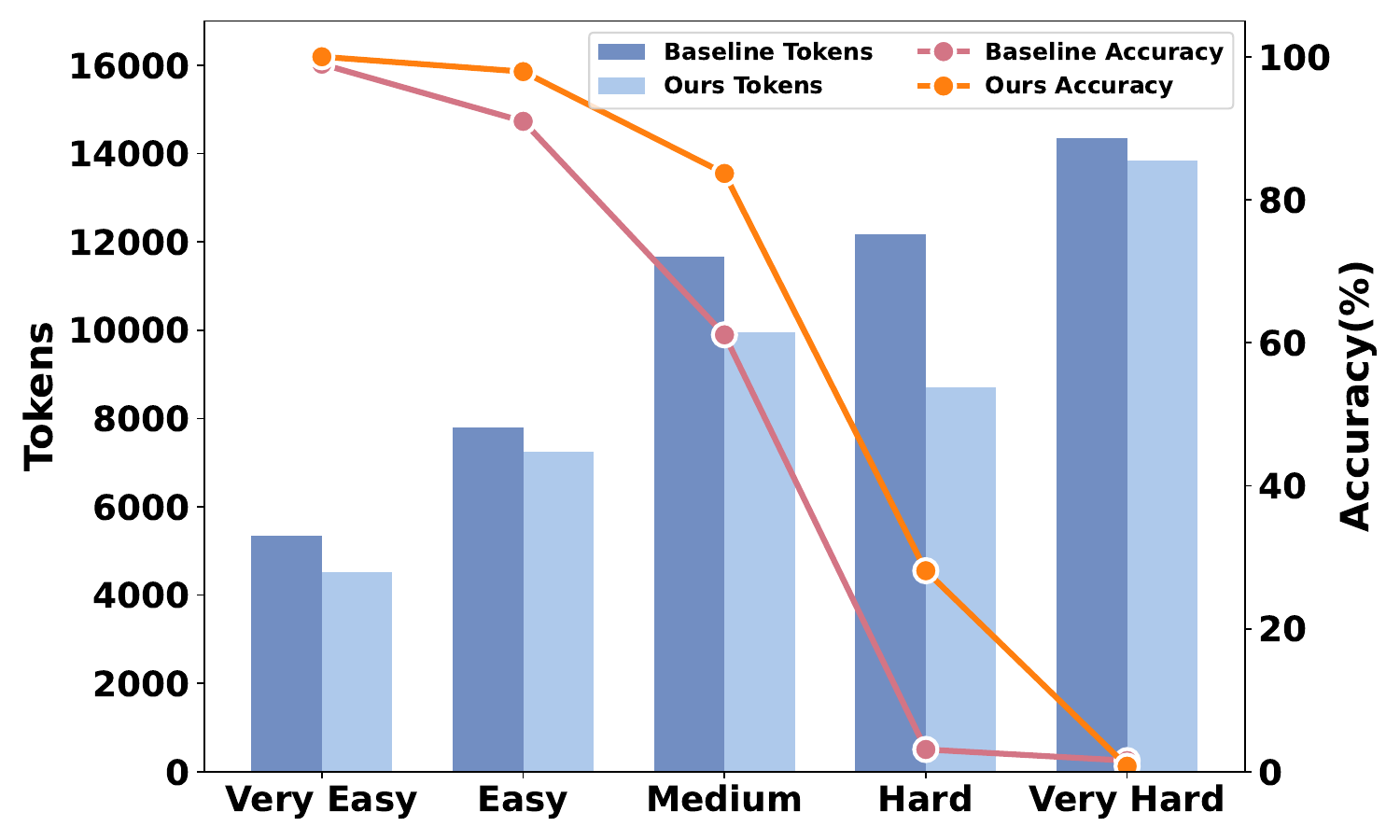}
    \caption{The Pass@1 and average token distribution across different difficulty levels, from Very Easy to Very Hard, for the LeaP-T-14B.}
    \label{fig:difficulty_leap_t_14B}
\end{figure}

We also report the performance of LeaP-T across the five difficulty levels of AIME 2024.
These difficulty levels follow the same categorization introduced in Section~\ref{sec:robustness_difficulty}, based on model~( without peer reasoning) accuracy.
To exclude the impact of distilling with a 32B model, we conduct distillation using SFT without LeaP, which serves as our baseline.
This setup allows for a direct comparison with our LeaP-T approach.
Across all five difficulty levels, LeaP-T generally outperforms the baseline.
For instance, as shown in Figure~\ref{fig:difficulty_leap_t_14B}, LeaP-T achieves higher accuracy while consuming fewer tokens.
These findings are consistent with the analysis presented in Section~\ref{sec:robustness_difficulty}.
This is because LeaP facilitates earlier consensus during the reasoning process, thereby reducing unnecessary computational overhead caused by overthinking.
Notably, the improvement of LeaP-T over the baseline suggests that its performance gain is not solely attributed to knowledge transfer through distillation.
Instead, it underscores the effectiveness of a training paradigm centered on \textit{learning from peers}.

\begin{figure}[H]
    \centering
    \includegraphics[width=.5\linewidth]{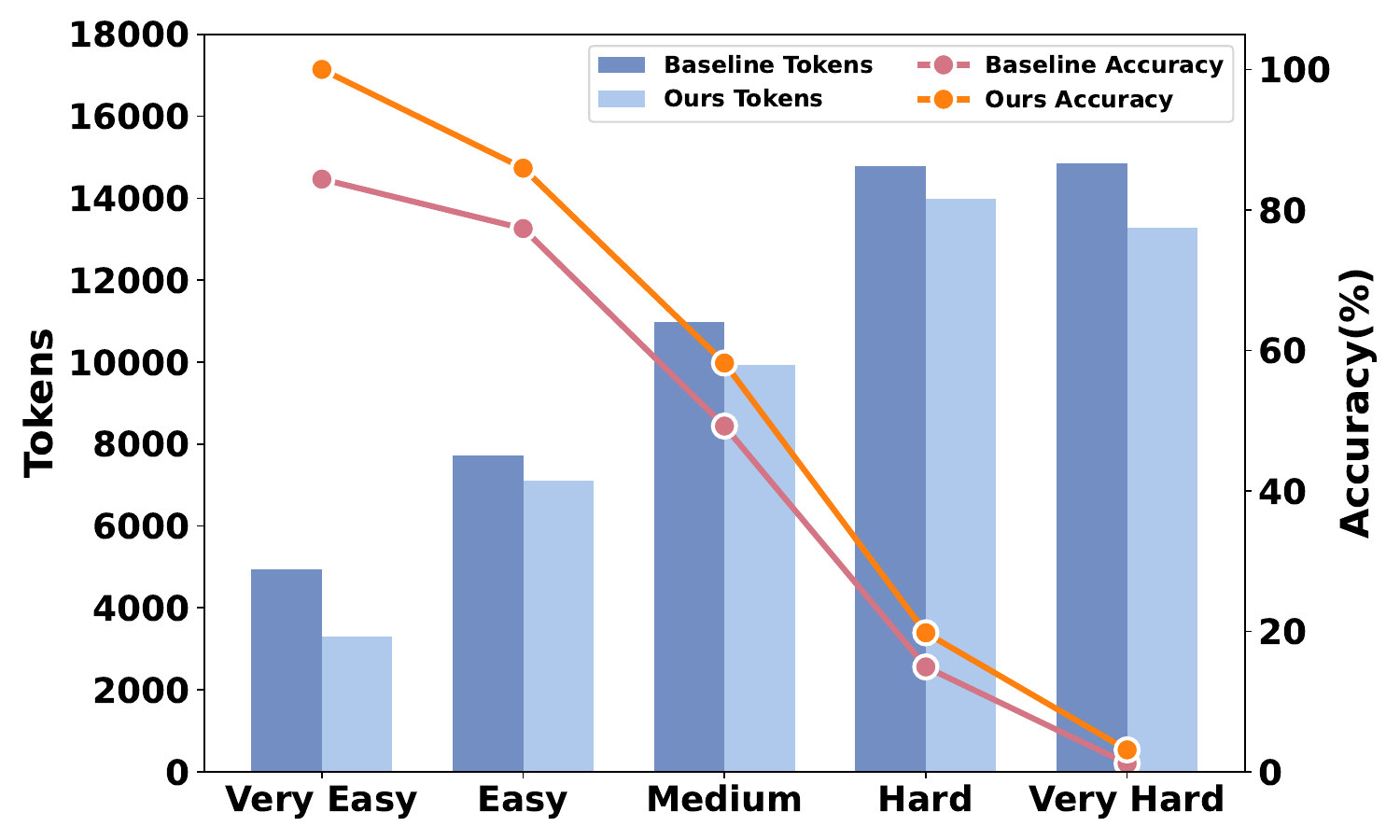}
    \caption{The Pass@1 and average token distribution across different difficulty levels, from Very Easy to Very Hard, for the LeaP-T-1.5B.}
    \label{fig:difficulty_leap_t_1.5B}
\end{figure}

\begin{figure}[H]
    \centering
    \includegraphics[width=.5\linewidth]{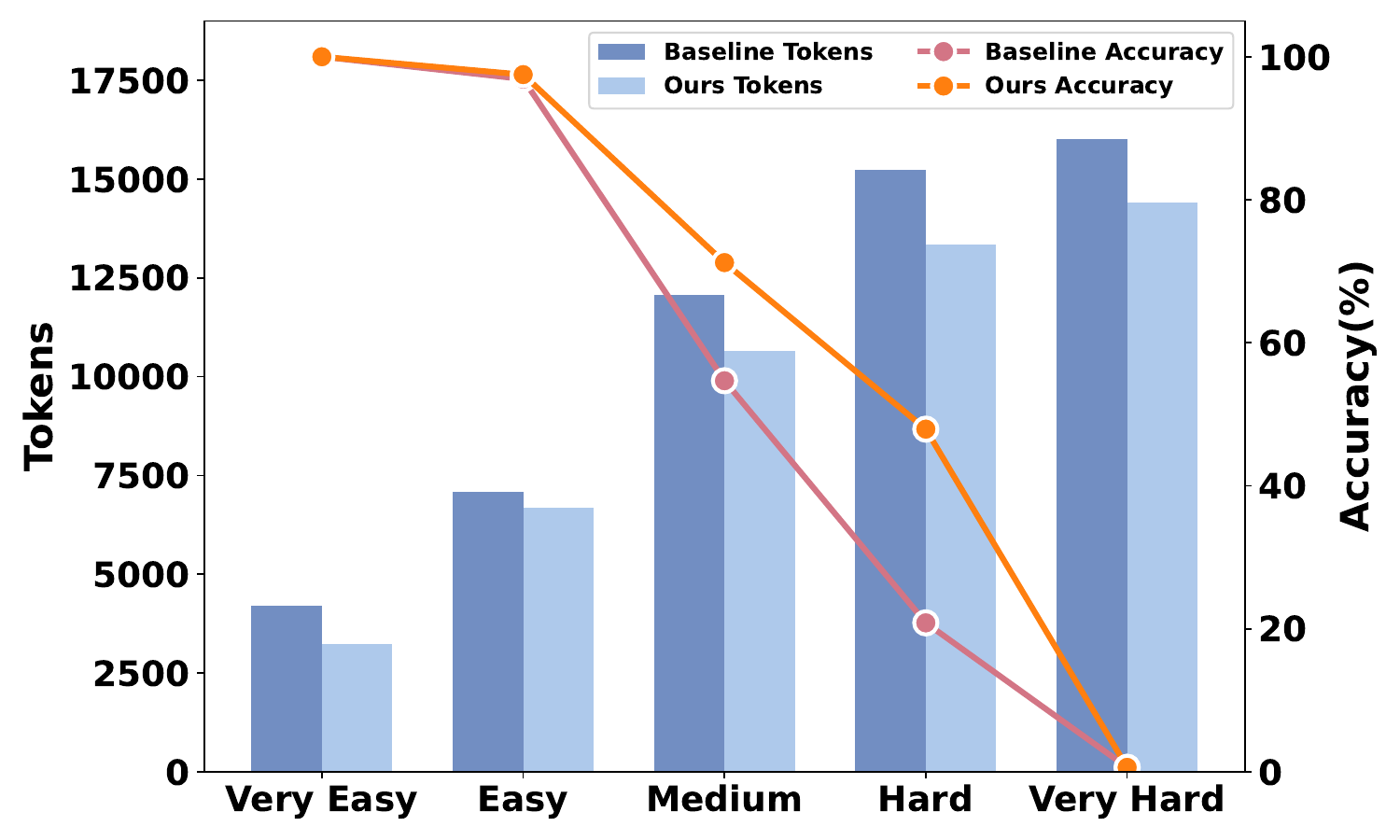}
    \caption{The Pass@1 and average token distribution across different difficulty levels, from Very Easy to Very Hard, for the LeaP-T-7B.}
    \label{fig:difficulty_leap_t_7B}
\end{figure}

\subsection{Behavior Difference of LeaP using RL-\textit{ed} and Non-RL models}
We observe that reinforcement learning (RL) models, such as QwQ, display higher confidence in their reasoning.
This is particularly evident in the increasing proportion of \textit{Unaffected} cases at later stages of reasoning.
The reinforcement learning process tends to encourage more consistent behavior, with certain tokens' sampling probabilities increasing, guiding the model back to familiar reasoning paths.
This aligns with findings from a previous study~\cite{ji2025first}, which shows that RL-trained models tend to prefer their learned reasoning strategies, exhibiting greater self-assurance in their conclusions. This confidence may explain the model’s resistance to altering its reasoning path when provided with peer insights, especially in the later stages of reasoning.

\end{document}